\documentclass[lettersize,journal]{IEEEtran}
\usepackage{amsmath,amsfonts}
\usepackage{algorithmic}
\usepackage{algorithm}
\usepackage{array}
\usepackage{textcomp}
\usepackage{stfloats}
\usepackage{url}
\usepackage{verbatim}
\usepackage{graphicx}
\usepackage{cite}
\usepackage{tikz}
\usepackage{soul}
\usepackage{tcolorbox}

\usetikzlibrary{tikzmark}
\usetikzlibrary{decorations.pathmorphing}
\usepackage{fancyhdr}

\newcommand{\redbox}[1]{%
  \tikz[baseline=(char.base)]{
    \node[anchor=text, rectangle, draw=red, dashed, thick, inner sep=2pt] (char) {#1};
  }%
}

\usepackage{caption}

\usepackage{hyperref}
\hypersetup{hidelinks} 

\usepackage{subcaption}
\usepackage{multirow}
\usepackage{makecell}
\usepackage{amssymb}

\usepackage{xcolor}

\usepackage{lipsum}
\usepackage{graphicx}

\newcommand{\xdownarrow}[1]{%
  {\left\downarrow\vbox to #1{}\right.\kern-\nulldelimiterspace}
}

\begin{document}

% \title{Learning a Universal Degradation Representation for Image Restoration}From Degraded to Restored: 
\title{Neural Degradation Representation Learning for All-In-One Image Restoration}

\author{Mingde Yao, Ruikang Xu, Yuanshen Guan, Jie Huang, and Zhiwei Xiong,~\IEEEmembership{Member,~IEEE}
        % <-this % stops a space

\thanks{
All authors are with the MoE Key Laboratory of Brain-inspired Intelligent Perception and Cognition, University of Science and Technology of China, Hefei, 230026, China.}}
% <-this % stops a space
% Corresponding author: Zhiwei Xiong.

\markboth{Journal of \LaTeX\ Class Files,~Vol.~14, No.~8, August~2021}%
{Shell \MakeLowercase{\textit{et al.}}: A Sample Article Using IEEEtran.cls for IEEE Journals}

\maketitle

\begin{abstract}

Existing methods have demonstrated effective performance on a single degradation type. In practical applications, however, the degradation is often unknown, and the mismatch between the model and the degradation will result in a severe performance drop. In this paper, we propose an all-in-one image restoration network that tackles multiple degradations. Due to the heterogeneous nature of different types of degradations, it is difficult to process multiple degradations in a single network. 
To this end, we propose to learn a neural degradation representation (NDR) that captures the underlying characteristics of various degradations. The learned NDR adaptively decomposes different types of degradations, similar to a neural dictionary that represents basic degradation components.
Subsequently, we develop a degradation query module and a degradation injection module to effectively approximate and utilize the specific degradation based on NDR, enabling the all-in-one restoration ability for multiple degradations. Moreover, we propose a bidirectional optimization strategy to effectively drive NDR to learn the degradation representation by optimizing the degradation and restoration processes alternately.
Comprehensive experiments on representative types of degradations (including noise, haze, rain, and downsampling) demonstrate the effectiveness and generalizability of our method. Code is available at \href{https://github.com/mdyao/NDR-Restore}{https://github.com/mdyao/NDR-Restore}.

\end{abstract}

\begin{IEEEkeywords}
All-in-one image restoration, degradation representation, denoising, deraining, dehazing, super-resolution
\end{IEEEkeywords}

\section{Introduction}

Image restoration aims to recover high-resolution and clean images from degraded or low-quality images, thereby improving visual quality and benefiting downstream applications. To process different types of degradations, various image restoration methods have been proposed, \emph{e.g.,} denoising~\cite{dabov2007color,zhang2017beyond,zhang2017learning,zhang2018ffdnet,tian2020image}, deraining~\cite{zhang2018density,yasarla2019uncertainty,wei2019semi,jiang2020multi,fu2019lightweight}, dehazing~\cite{cai2016dehazenet,ren2016single,li2017aod,qu2019enhanced,dong2020fd}, and super-resolution (SR)~\cite{dong2015image,lim2017enhanced,xu2023zero,niu2020single,kim2016accurate}. These methods have shown great potential in addressing various image restoration tasks, which facilitates their application in practice.

\begin{figure}[t]
  \centering
  \includegraphics[width=\linewidth]{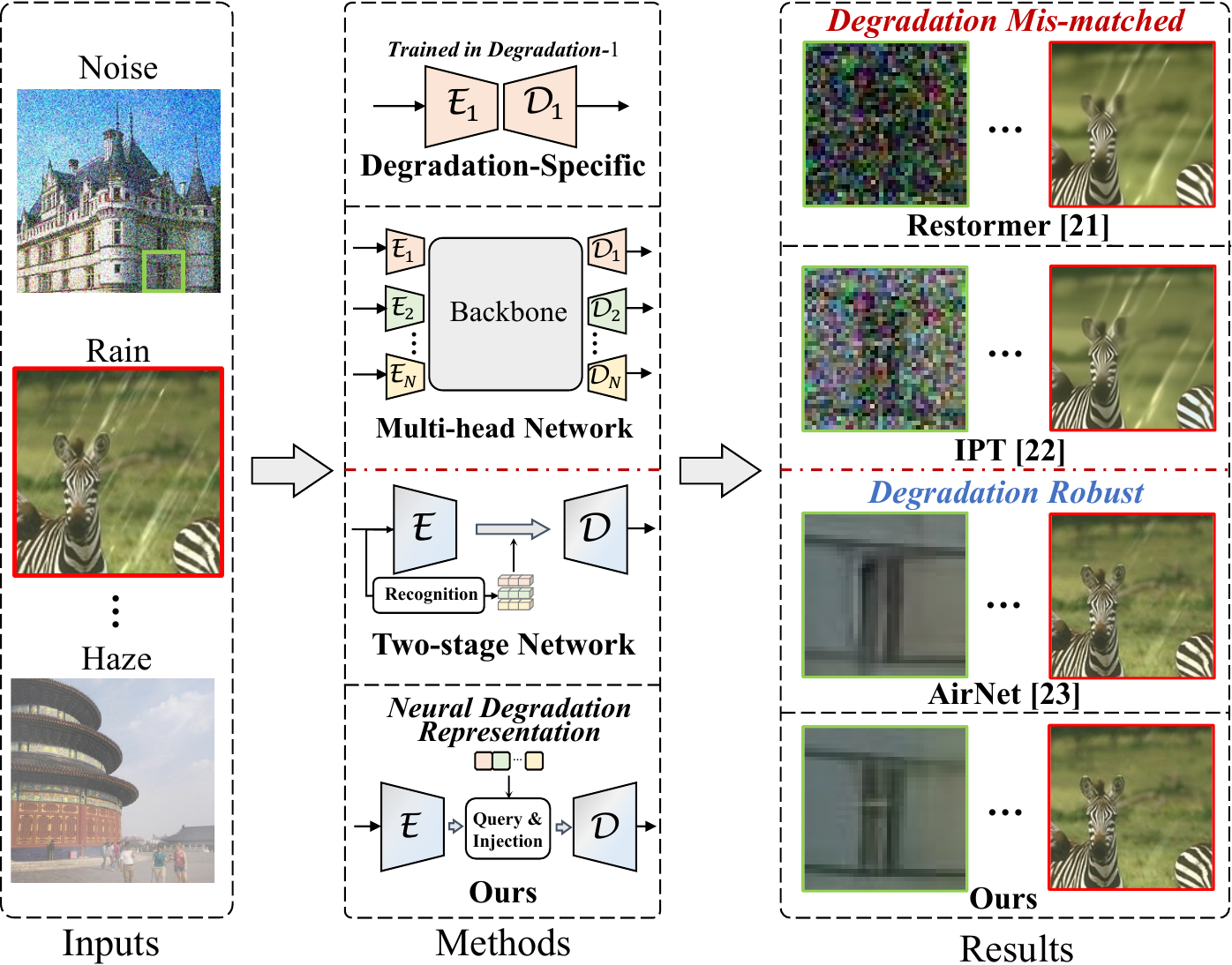}
  \caption{Comparison between our method and other methods. \cite{zamir2022restormer,chen2021pre} fail to restore clean image if the model mismatches the degradation. 
  Our method handles multiple degradations with a single network and produces more visually appealing results than the existing two-stage all-in-one model~\cite{li2022all}.}
  \label{fig:1_intro_teaser}
  \vspace{-0.6cm}
\end{figure}

However, existing restoration methods are generally limited to one type of degradation and cannot be readily applied to multiple degradations, posing a challenging task in real-world applications. This is because the image degradation present in real-world scenarios is often unknown ({as per \cite{li2022all}, we use the term ``unknown'' to describe unspecific degradation, and it should not be confused with ``unseen'' degradation}), and using a mismatched model for a specific type of degradation can result in a significant performance drop, as shown in Fig.~\ref{fig:1_intro_teaser}. An alternative approach is to first assess the type of degradation and then select an appropriate model from a model library for restoration. However, this approach requires a large model library and additional degradation assessment procedures, resulting in increased storage and computational overhead. Another approach, similar to the image signal processing (ISP) pipeline, is to sequentially apply all possible restoration models to restore the degraded image, but it still suffers from the problem of computational redundancy and error accumulation.

Therefore, it is imperative to develop an all-in-one image restoration model that can handle various degradations using a single network. However, this is challenging due to the complex mapping relationships~\cite{wang2022relationship} between various degraded inputs and clean outputs, resulting in a network that is difficult to optimize. In addition, different degradations may possess distinct statistical properties that interfere with each other~\cite{li2022all,wang2021unsupervised}, leading to a decrease in performance. These difficulties hinder the effective handling of multiple degradations in a single network.

Recent studies have explored the feasibility of image restoration of multiple degradations, as shown in Fig.~\ref{fig:1_intro_teaser}. IPT~\cite{chen2021pre} employs multiple heads and tails with a shared body to process different types of degradations. However, the redundant heads and tails in IPT cause deployment challenges, and it still relies on prior knowledge for head/tail selection. AirNet~\cite{li2022all} proposes an all-in-one image restoration network that utilizes contrastive learning to distinguish different degradations. By treating similar degradations as positive pairs and different degradations as negative pairs, it acquires distinct degradation representations for subsequent image restoration. However, AirNet requires two-stage training and additional training costs to support contrastive learning.

Unlike previous methods~\cite{chen2021pre,li2022all}, in this paper, we propose learning a neural degradation representation (NDR) that effectively captures the essential characteristics of various degradations. {NDR is a learnable tensor, initialized randomly and optimized adaptively through the training process. }
% Each vector in NDR serves as a type of learned degradation, which means it is not specific to the handcrafted degradation.} 
By leveraging NDR, our proposed all-in-one image restoration network, named NDR-Restore, can approximate and utilize the specific degradation of the input image, enabling adaptive restoration within a single network. 
Unlike existing representation learning methods~\cite{zhou2022joint,wang2021unsupervised} that focus on capturing texture and content details, NDR is specifically designed to learn the degradation representation. {This enables training NDR-Restore in an end-to-end way and avoids the complexity of constructing positive/negative pairs~\cite{li2022all}. }

To build NDR-Restore, we propose a degradation query (DQ) module for degradation approximation and a degradation injection (DI) module for degradation utilization, allowing NDR-Restore to handle multiple degradations.
Specifically, the DQ module is designed to query the degradation representation from NDR, which plays a key role in image restoration. This process yields a degradation tensor that contains degradation information of the input image. 
Then, the DI module injects the degradation tensor into the image feature for image restoration. In the DI module, we introduce low-rank feature modulation to project the degradation tensor and the image feature into the same space, facilitating the interaction between degradation information and image features. Finally, we seamlessly integrate the DQ and DI modules into a hierarchical encoder-decoder architecture to achieve robust restoration for different degradations.

For network optimization, we propose a bidirectional optimization strategy. 
Specifically, we introduce NDR-Degrad, an auxiliary degradation network that is jointly optimized with NDR-Restore. During the training process, NDR-Restore generates a clean image and queries the degradation tensor from NDR, while NDR-Degrad utilizes the queried degradation tensor to degrade a clean image. 
By optimizing NDR-Restore and NDR-Degrad alternately, we implicitly drive NDR to learn the degradation representation. 
This strategy relies on the rationale that, if NDR-Degrad could generate the specific degraded image, it indicates the queried degradation tensor effectively captures the degradation information from NDR. In other words, NDR indeed represents degradation.
The NDR-Degrad is only utilized for training, and once trained, we only need NDR-Restore for restoration.

We conducted comprehensive experiments on representative image restoration tasks, including denoising, deraining, dehazing, and SR. The experimental results demonstrate that NDR-Restore can effectively handle multiple degradations and outperforms existing all-in-one image restoration methods. Moreover, we evaluate our approach on real-captured images, revealing its great potential for applications in real-world scenarios. In summary, our contributions are as follows.

\begin{itemize}

    \item We propose a novel method for all-in-one image restoration, which provides a practical solution for handling multiple degradations using a single network. 

    \item We propose NDR to represent underlying characteristics and statistical properties of various degradations. Based on NDR, we devise two novel modules, DQ and DI, to effectively approximate and utilize the specific degradation during image restoration. 

    \item {We propose a bidirectional optimization strategy that imposes additional constraints within the restoration network, thereby enhancing the overall performance.}

    \item We conduct comprehensive experiments on a number of representative image restoration tasks to demonstrate the superiority of NDR-Restore over existing methods. 
\end{itemize}

\vspace{-0.3cm}
\section{Related Work}
\vspace{-0.1cm}
\subsection{Image Restoration}
Image restoration is a fundamental task in computer vision, aiming to recover the original high-quality image from its degraded or corrupted version. In recent years, researchers have proposed tremendous neural networks that are tailored to specific tasks, such as denoising~\cite{zhang2017beyond,zhang2018ffdnet,chen2018deep,yao2023towards}, deraining~\cite{li2018recurrent,jiang2020multi,li2018recurrent,wang2020model,wei2019semi,yasarla2019uncertainty,zhang2018density,xiao2022image,zhang2023data,wei2021deraincyclegan}, dehazing~\cite{cai2016dehazenet,he2010single,li2021you,liu2018learning}, and SR~\cite{zhang2018residual,wang2018esrgan,chen2023activating,pan2022towards}. 
For example, DnCNN~\cite{zhang2017beyond} utilizes a deep convolutional architecture to learn the mapping between noisy and clean images, effectively suppressing the noise while preserving image details. 
IDT~\cite{xiao2022image} utilizes a Transformer-based~\cite{vaswani2017attention} architecture to capture the long-range dependencies for image deraining. N2V~\cite{krull2019noise2void} proposes an self-supervised learning-based method with blind-spot convolution. 
Although the aforementioned methods have made significant progress, they are still confined to handling single degradations, which restricts their broad applicability.

All-in-one image restoration, which utilizes a single network to handle different restoration tasks, has emerged as a promising direction~\cite{zhang2023all,li2022all,potlapalli2023promptir,he2020momentum}. Early attempts~\cite{zamir2022restormer,mehri2021mprnet} utilize the same network architecture trained on different tasks with different parameters. However, training and deploying such networks for each task can still be inefficient. To simplify the network, IPT~\cite{chen2021pre} introduces body-sharing and uses different heads and tails for different restoration tasks. However, it still requires recognizing the degradation with additional degradation assessment and lacks versatility. More recently, AirNet~\cite{li2022all} proposes an all-in-one image restoration network by leveraging contrastive learning~\cite{he2020momentum}. ADMS~\cite{park2023all} utilizes adaptive discriminative filters to handle different degradations and IDR~\cite{zhang2023ingredient} proposes a two-stage ingredients-oriented restoration network. Meanwhile, unified weather restoration methods~\cite{li2020all,valanarasu2022transweather,chen2022learning} are also proposed. {PromptIR~\cite{potlapalli2023promptir} has recently been proposed utilizing learnable degradation-related parameters. However, these methods are limited by degradation representations. In contrast, we utilize a more reasonable architecture for incorporating degradation information at the pixel level, providing a more effective and versatile approach to all-in-one image restoration.}

 % and require large efforts of two-stage training processes (or multi-head architectures). 
% nd we our method adopts an end-to-end learning strategy to train NDR with different degradations mixed, providing a more effective and versatile approach to all-in-one image restoration.}

\begin{figure*}[t]
  \centering
  \includegraphics[width=1\linewidth]{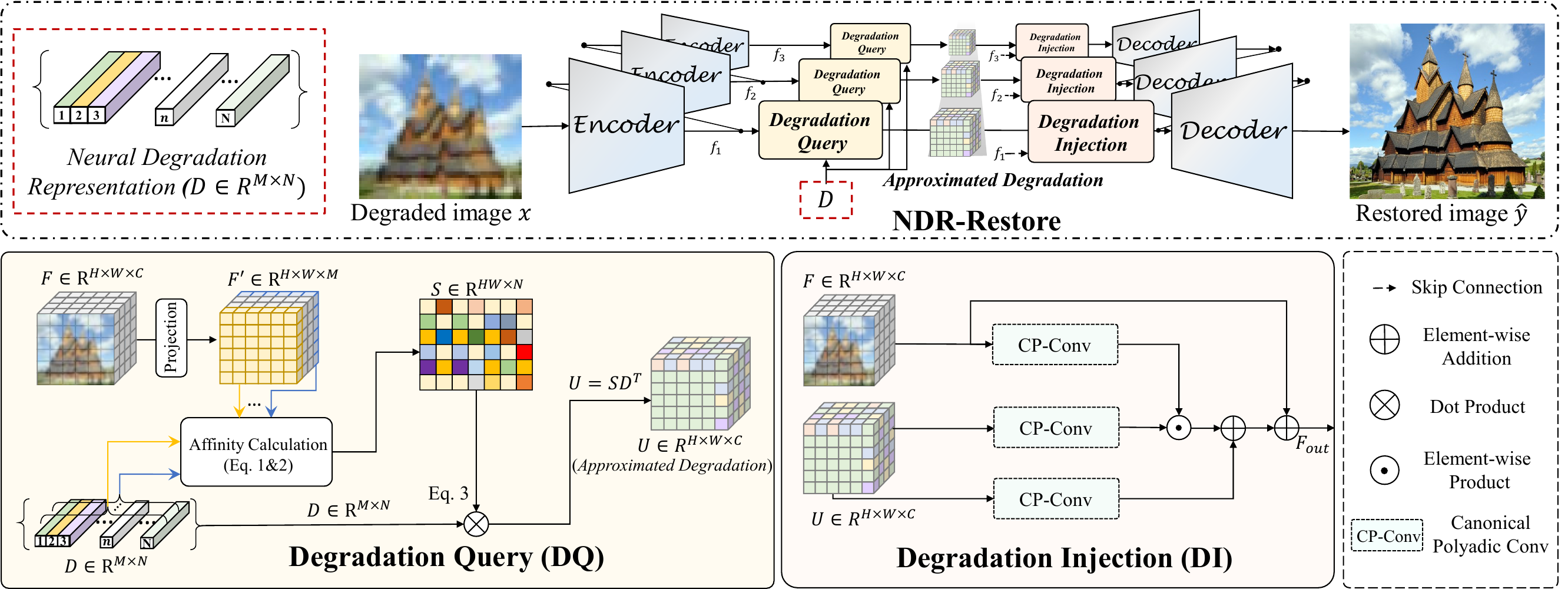}
  \caption{Overview of our method. We construct NDR-Restore using a multi-scale architecture. NDR-Restore utilizes the DQ module to approach degradation and leverages the DI module to facilitate the interaction between degradation information and image features. NDR captures the underlying characteristics of multiple degradations and is utilized in the DQ module to generate degradation.  }
  \label{fig:3pipeline}
  \vspace{-0.5cm}
\end{figure*}

\vspace{-0.3cm}
\subsection{Degradation Representation Learning}

The conventional image restoration pipeline uses a predefined model to represent degradation, \emph{e.g.,} Gaussian noise~\cite{zhang2017beyond,zheng2021deep} or motion blur~\cite{pan2016blind,ren2016image}. The restoration model is trained to reverse the degradation process and restore the clean image based on the provided degradation model. In real-world scenarios, however, the degradation process is often unknown and complex, and explicit degradation representations become limited~\cite{liu2022tape}. Hence, it is crucial to find appropriate degradation representations for all-in-one image restoration.

To overcome this challenge, recent works use neural networks to learn degradation representations.	
DAN~\cite{luo2020unfolding} uses an unfolding algorithm to learn a degradation kernel, capturing the intrinsic features of spatial resolution degradation.
DASR~\cite{wang2021unsupervised} presents an unsupervised scheme to learn representations between various degradations in the feature space, enabling the network to adapt flexibly to different degradations in blind SR tasks.
Similarly, AirNet~\cite{li2022all} proposes a contrastive learning-based approach to learn a degradation representation for all-in-one image restoration. They train the network by treating similar degradations as positive pairs and different degradations as negative pairs, resulting in distinguishable degradation representations. {In contrast to these methods, our proposed approach adaptively and directly learns a wide range of degradation characteristics from the data, reducing the reliance on manually predefined categories.}

\vspace{-0.2cm}
\section{Method}
\vspace{-0.2cm}
\subsection{Overview}

We propose neural degradation representation (NDR) to effectively represent the intrinsic features of different degradations, enabling a single network to handle multiple types of degradation. {Neural Degradation Representation (NDR) is a learnable tensor that represents the intrinsic features of different degradations, which is randomly initialized and adaptively learned from the restoration and degradation processes. Specifically, NDR can be denoted as $D \in \mathbb{R}^{M \times N}$, where $N$ is the number of degradation types and $M$ is the feature dimension of each degradation. Each vector in NDR serves as a type of learned degradation, which means it is not specific to hand-crafted degradation. NDR is independent of the context information and is utilized in a query mechanism for degradation approximation (see Sec.~\ref{sec:DQ}).}

\begin{figure*}[t]
  \centering
  \includegraphics[width=0.98\linewidth]{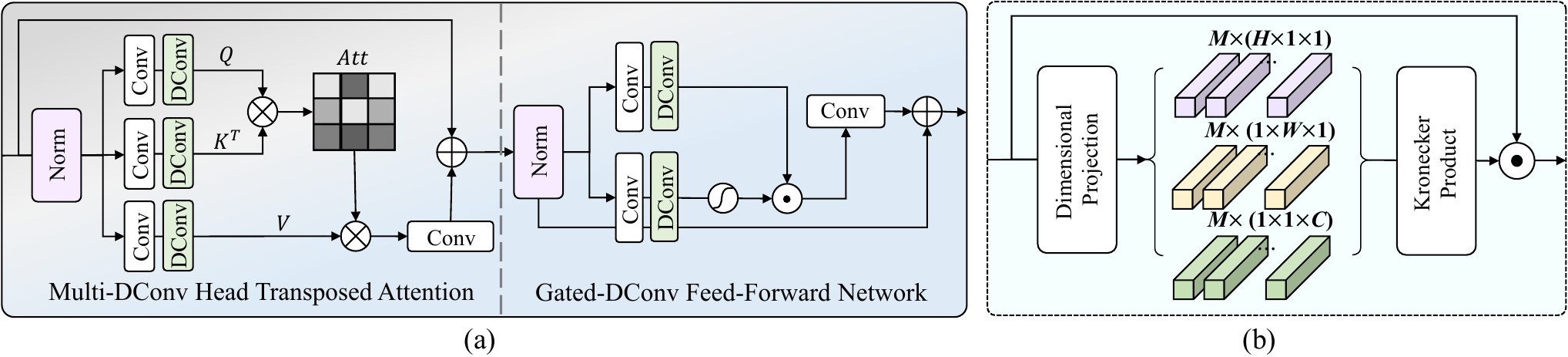}
  \vspace{-0.2cm}
  \caption{Network details. (a) The implementation of the encoder and decoder in NDR-Restore. It takes transformer-based attention mechanism~\cite{zamir2022restormer} to extract shallow and deep features. (b) The details of CP-Conv, referred to Eq.~\ref{eq:cp1}~\&~\ref{eq:cp2}. }
  \label{fig:3detail}
  \vspace{-0.5cm}
\end{figure*}

To leverage NDR, we construct an all-in-one image restoration network NDR-Restore in a multi-scale encoder-decoder structure. In NDR-Restore, we introduce two novel modules for effective NDR utilization: the degradation query (DQ) module, which facilitates approximation of the image's degradation, and the degradation injection (DI) module, which allows for the utilization of the image's degradation into the restoration process, thereby enabling the all-in-one image restoration.

The pipeline of NDR-Restore is shown in Fig.~\ref{fig:3pipeline}. Given a degraded image $x\in \mathbb{R}\ ^{H\times W \times 3}$, it is first fed into the encoder to extract the deep feature $F\in \mathbb{R}\ ^{H\times W \times C}$, where $H, W, C$ represent the height, width, and channel shape, respectively. The feature $F$ contains both the context and degradation information of the current image. Then, the DQ module takes the $F$ and the degradation representation $D$ to obtain the approximated degradation $U\in \mathbb{R}^{H\times W \times C}$, which represents the degradation of the current image and is fine-grained for each pixel. Subsequently, the DI module injects the approximated degradation $U$ to the image feature $F$, enabling the degradation utilization with context information. Finally, the features are sent into the decoder to reconstruct a clean image $\hat{y}\in\mathbb{R}^{H\times W\times 3}$. It is worth noting that, we take the original scale inside the multi-scale architecture for illustration, while other scales perform in a similar way. We take transformer-based attention mechanism~\cite{zamir2022restormer} to implement the encoder and decoder, as shown in Fig.~\ref{fig:3detail}(a).

To optimize NDR, we propose a bidirectional optimization strategy by introducing a degradation network NDR-Degrad. During the training process, we optimize both directions (\emph{i.e.,} NDR-Degrad and NDR-Restore) to drive NDR to represent degradation. Specifically, NDR-Restore inputs the degraded image and predicts a clean image as well as an approximated degradation. The NDR-Degrad takes the approximated degradation and a clean image to generate a degraded image. The overall optimization objective is to minimize the loss in both NDR-Restore and NDR-Degrad. Since the approximated degradation is queried from NDR, optimizing NDR-Degrad can implicitly drive the degradation learning of NDR, such that NDR is forced to capture the essential features of various image degradations. During the inference process, we only require NDR-Restore for all-in-one image restoration.

\begin{figure*}[t]
    \centering
    \includegraphics[width=1\linewidth]{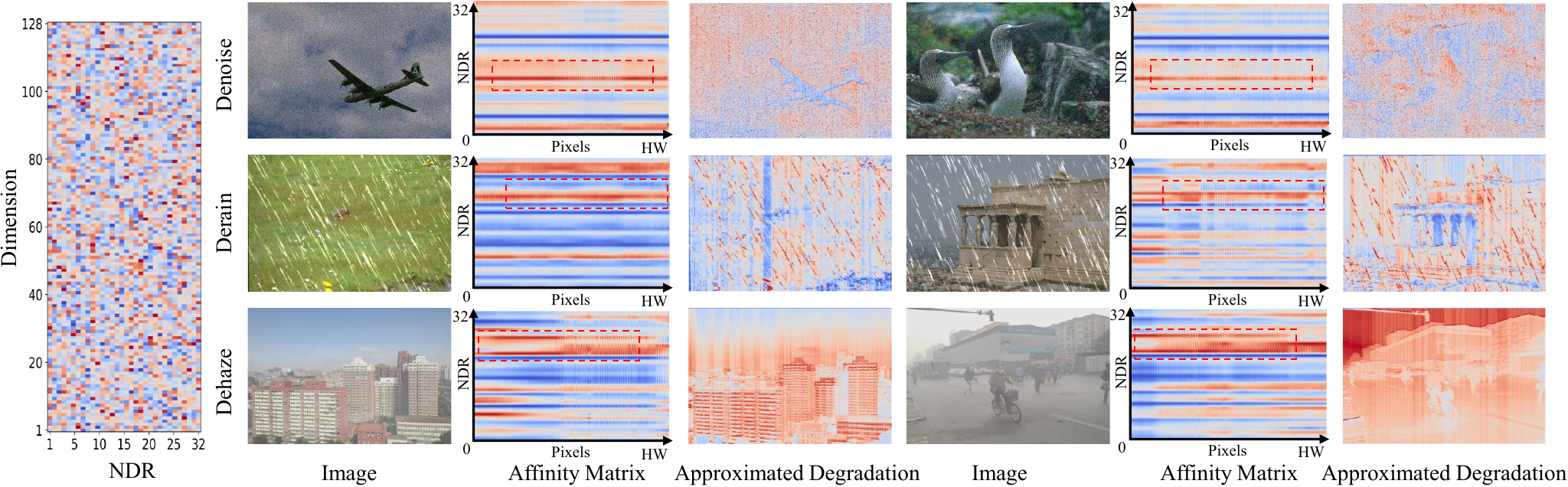}
    \vspace{-0.4cm}
    \caption{Visualizations of NDR $D$, affinity matrices $S$, and approximated degradations $U$. Please note \protect\redbox{red dashed rectangles} in $S$. We can observe distinguishing activation of different degradations and similar activation of the same degradation, which demonstrates the effective degradation approximation of DQ module and degradation representation of NDR. }

    \label{fig:get_M}
    \vspace{-0.5cm}
\end{figure*}

\vspace{-0.2cm}
\subsection{Degradation Query}\label{sec:DQ}

To facilitate the approximation of degradation in the current image, we propose the DQ module, which queries the degradation from NDR. This module generates the approximated degradation $U$ that represents the degradation of the degraded image. The DQ module consists of three main parts: feature mapping, affinity calculation, and degradation query, as shown in Fig.~\ref{fig:3pipeline}.

{First, we map the input feature map $F$ to the same channel dimension, using 1$\times$1 convolution layers. Subsequently, we flatten the mapped $F^{'} \in \mathbb{R}^{H\times W \times M}$ into a two-dimensional tensor $P \in \mathbb{R}^{HW \times M}$. }

{Then we calculate the affinity, which quantifies the relationship between $hw$-th feature and $n$-th degradation. For $hw$-th feature $P_{hw}\in\mathbb{R}^{1\times M}$, we perform dot product with $n$-th degradation $D_{n}\in \mathbb{R}^{M\times 1}$, resulting in the affinity scalar as} 
\vspace{-0.3cm}
\begin{equation}\label{eq:get_sigma}
\sigma_{hw,n} = \sum_{m=1}^{M} P_{hw,m} \cdot D_{m,n},
\vspace{-0.1cm}
\end{equation}
{where $\sigma_{hw,n}$ represents the $n$-th affinity scalar between the $hw$-th feature and the $n$-th degradation. In other words, there are total $N$ affinity scalars for each pixel.
Each $\sigma$ quantifies how much the degradation affects the pixel.}

{For stabilizing the training process, we adopt a softmax operation to normalize $\sigma_{hw,n}$, ensuring that the sum of contributions from $N$ degradations is $1$. 
This 
yields an affinity matrix $S\in \mathbb{R}^{HW\times N}$, which can be expressed as}
\vspace{-0.1cm}
\begin{equation}\label{eq:get_S}
S_{hw,n} = \frac{e^{\sigma_{hw,n}}}{\sum_{n=1}^{N} e^{\sigma_{hw,n}}}.
\vspace{-0.1cm}
\end{equation}
{We discuss the differences between $1\times1$ convolution and Eq.~\ref{eq:get_sigma} in Sec.~\ref{sec:4_11conv}.}

Finally, we query the degradation information from $D$, thereby approximating the degradation $U\in \mathbb{R}^{H\times W\times C_{in}}$ for the current image. {In this step, the affinity matrix $S$ is leveraged to re-weight $D$ \textbf{along the $N$ dimension} of NDR.} As a result, we obtain the approximated degradation $U$ that captures the degradation information of the input image.
To be detailed, we first utilize $S$ to query the degradation from $D$ as
\vspace{-0.2cm}
\begin{equation}\label{eq:get_U}
U'_{hw,m} = \sum_{n=1}^{N} S_{hw,n} \cdot D^T_{n,m},
\vspace{-0.15cm}
\end{equation}
where $U'_{hw,m}$ represents the approximated degradation value at pixel location $(h, w)$ with $m$-th dimension. {We transpose $D$ to $D^T$ in Eq.~\ref{eq:get_U} to reweight different degradations along the $N$ dimension.} Subsequently, we reshape $U' \in \mathbb{R}^{HW\times M}$ back to $\mathbb{R}^{H\times W \times M}$ and employ a 1$\times$1 convolution layer to map it to $U\in\mathbb{R}^{H\times W \times C}$. Consequently, $U$ becomes a degradation tensor where each pixel corresponds to a fine-grained approximated degradation. It can be effectively injected into the following image restoration process to adaptively remove degradation and restore clean images.

In Fig.~\ref{fig:get_M}, we present visualizations of the affinity matrix $S$ and the approximated degradation $U$. For affinity matrix $S$, similar activations are observed along the same degradation type. This observation highlights the effectiveness of NDR in capturing the characteristic features of various degradations, while the DQ module successfully establishes corresponding relationships between NDR and the degraded image feature $F$. Moreover, the visualization of $U$ reveals the spatial alignment of degradation patterns with the image pixels, which demonstrates the DQ module can approximating and represent the specific degradation of the current image.

\begin{figure}[t]
    \centering
    \includegraphics[width=1\linewidth]{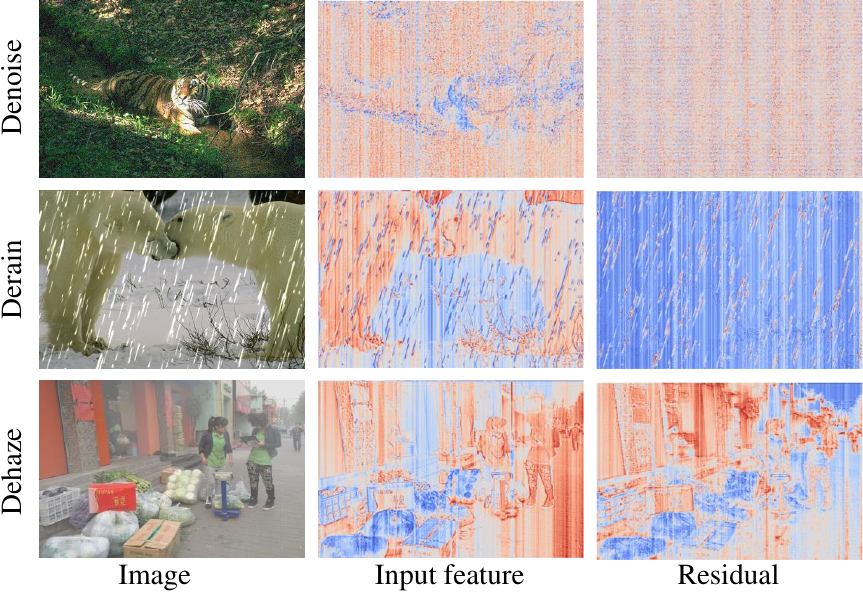}
    \caption{Features processed by the DI module. The residual features (before and after the DI module) only contain the degradation information, which demonstrates the effective degradation removal of the DI module.}
    \label{fig:3DI_module}
    \vspace{-0.5cm}
\end{figure}

\vspace{-0.2cm}
\subsection{Degradation Injection}\label{sec:3DI_module}

We propose the DI module to effectively utilize the approximated degradation $U$ for adaptive degradation removal and clean image restoration. There are two challenges to be addressed. First, processing distinct degradation and image information is crucial, as they exist in different spaces. Second, since the degradation information is spatially related, the degradation utilization should consider the pixel-wise and content-aware nature of the image, thus ensuring the injected degradation aligns with the image content.

To overcome these challenges, DI module maps the degradation and image features to the same space, allowing spatial utilization of the approximated degradation. We show the pipeline of the DI module in Fig.~\ref{fig:3pipeline} and the details in Fig.~\ref{fig:3detail}(b). To achieve the mapping, we devise a Canonical Polyadic (CP)-Conv in the DI module, which can be written as
\vspace{-0.1cm}
\begin{equation}\label{eq:cp}
    U_{cp} = CP(U),\quad F_{cp}=CP(F). 
\end{equation}
% \vspace{-0.1cm}
Here, $CP$ represents the CP-Conv built on the CP decomposition~\cite{kolda2009tensor}, which effectively extracts the main characteristic and essential representation from the input data. Different from previous works~\cite{hu2022many} that directly apply CP decomposition to a single feature, we leverage it to simultaneously capture the essential representation of both image and degradation features, thereby mapping them into the same space. 

The CP-Conv consists of a dimension projection and a Kronecker product~\cite{van2000ubiquitous,yu2022efficient}. We take $U_{cp}=CP(U)$ in Eq.~\ref{eq:cp} as an example for simplicity, while $F_{cp}=CP(F)$ can be easily inferred. The dimension projection utilizes three learnable projectors ($p_1$, $p_2$, and $p_3$) to map the input data to sub-features, yielding three 1-D features:
\vspace{-0.1cm}
\begin{equation}\label{eq:cp1}
    U_1=p_1(U),\quad U_2=p_2(U),\quad U_3=p_3(U).
    \vspace{-0.1cm}
\end{equation}
Each projector comprises an average pooling layer, a 1$\times$1 convolutional layer, and a sigmoid function. Here, $U_1\in \mathbb{R}^{K \times (C \times 1 \times 1)},
    U_2\in \mathbb{R}^{K\times (1 \times H \times 1)},
    U_3\in \mathbb{R}^{K \times (1 \times 1 \times W)}
$, and $K$ is set to be less than $C$, $H$, and $W$.
These projections effectively integrate information along different dimensions, ensuring the low-rank property of the resulting features. 
Subsequently, the three features are conducted in the Kronecker product as
\vspace{-0.1cm}
\begin{equation}\label{eq:cp2}
    U_{Kro} = U_1\otimes U_2 \otimes U_3, U_{Kro}\in \mathbb{R}^{K\times H\times W\times C},
\end{equation}
where $\otimes$ denotes the Kronecker product that multiplies each element of a matrix with another complete matrix. 
Next, the $U_{Kro}$ is point-wise averaged along the $K$ dimension to obtain $U_{cp}$. Since $U_{cp}$ is derived from three rank-1 features, it has the low-rank property to lie in the closer space with the image feature $F_{cp}$, resulting in the effective degradation injection.

To ensure the injected
degradation aligns with the image content, we introduce the affine mapping as 
\vspace{-0.1cm}
\begin{equation}
   F_{out}=(F_{cp}\odot U_{cp} + U_{cp})+F,
\end{equation}
where $\odot$ denotes the element-wise production that spatially aligns the degradation with the image content. $F_{out}\in \mathbb{R}^{H\times W \times C}$ represents the output of the DI module. In Fig.~\ref{fig:3DI_module}, we show the input image feature $F$ and the corresponding residual map, which demonstrate the differences before and after the degradation injection.

\vspace{-0.2cm}
\subsection{Bidirectional Optimization}

\begin{figure}[t]
  \centering
  \includegraphics[width=1\linewidth]{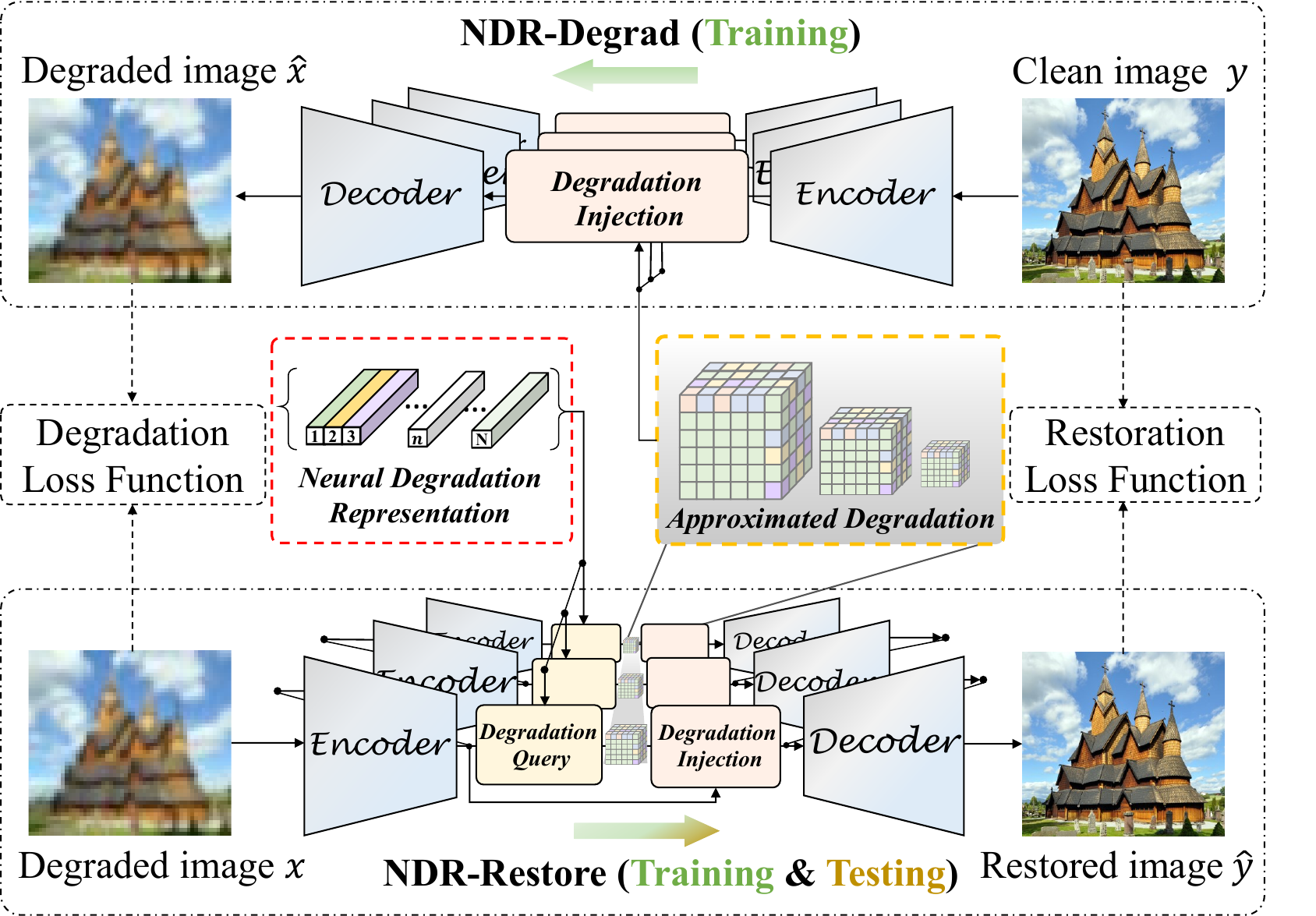}
  \caption{Overview of training strategy. The proposed strategy optimizes both degradation and restoration directions, by introducing an auxiliary degradation network NDR-Degrad, to drive the learning of NDR.}
  \label{fig:3train}
  \vspace{-0.5cm}
\end{figure}

As aforementioned, NDR plays a crucial role in representing degradations and is a key component in NDR-Restore. However, NDR is merely a set of learnable parameters and training NDR-Restore in an end-to-end manner alone may not provide NDR with a clear physical interpretation. Therefore, we propose a bidirectional training strategy that constrains the degradation and restoration processes to drive the NDR to represent degradation, as shown in Fig.~\ref{fig:3train}.

To this end, we introduce an auxiliary degradation network NDR-Degrad. Unlike the restoration network NDR-Restore, NDR-Degrad generates degraded images from clean images using the given degradation. {NDR-Degrad consists of an encoder, a DI module (see Sec.~\ref{sec:3DI_module}), and a decoder. The encoder and decoder use more lightweight architectures than in NDR-Restore by reducing the number of layers, as it provides assist supervision for the learning of NDR-Restore. The DI module injects the degradation $U$ into the extracted features, where $U$ is given by the DQ module from NDR-Restore.}

As shown in Fig.~\ref{fig:3train}, the bidirectional training strategy optimizes both NDR-Restore and NDR-Degrad. For NDR-Restore, we input the degraded image $x$ to obtain the restored image $y'$ and the approximated degradation $U$. For NDR-Degrad, we input the clean image $y$ and $U$ to generate a degraded image $x'$. We bidirectionally optimize the two networks using the following loss function
\vspace{-0.1cm}
\begin{equation}\label{eq:3bidirectional}
\mathcal{L}=\mathcal{L}_{restore}+\mathcal{L}_{degrad} =\|x-x'\|_2+\lambda \|y-y'\|_2,
\end{equation}
% \vspace{-0.1cm}
where $\lambda$ is a scaling factor.

The rationale behind this strategy is that if NDR-Degrad could generate a specific degraded image conditioned on the degradation tensor $U$, which is queried from NDR, it signifies that the queried tensor $U$ effectively captures the degradation information, thereby validating NDR as an effective representation of degradation. Consequently, during the bidirectional optimization process, NDR is implicitly driven to learn degradation representations. During the inference process, we only require NDR-Restore for all-in-one image restoration without the auxiliary network NDR-Degrad.

\vspace{-0.2cm}
\section{Experiments}
% \vspace{-0.2cm}

We conduct experiments on two degradation settings,\textit{ i.e.}, single degradation and multiple degradations. It is worth noting that, we do NOT aim to achieve state-of-the-art (SOTA) performance on single degradation since our focus is not the architecture/algorithm design on a specific task. Despite this, we report experimental results on the single degradation to demonstrate: 1) our method can work well on the single degradation, and 2) we provide an anchor to better analyze the performance on multiple degradations.

\begin{table*}[t]
\centering
\vspace{-0.5cm}
\caption{Quantitative results of denoising on the BSD68~\cite{martin2001database} and Urban100~\cite{huang2015single} datasets.}
\label{tab:4denoising}
\vspace{-0.2cm}
% \resizebox{\textwidth}{!}{%
\begin{tabular}{c|ccc|ccc}
\hline
\multirow{2}{*}{Method} & \multicolumn{3}{c|}{BSD68~\cite{martin2001database}} & \multicolumn{3}{c}{Urban100~\cite{huang2015single}} \\
 & $\sigma$=15 & $\sigma$=25 & \multicolumn{1}{c|}{$\sigma$=50} & $\sigma$=15 & $\sigma$=25 & $\sigma$=50 \\ \hline
CBM3D~\cite{dabov2007color} & 33.50/0.9215 & 30.69/0.8672 & {27.36/0.7626} & 33.93/0.9408 & 31.36/0.9092 & 27.93/0.8404 \\
DnCNN~\cite{zhang2017beyond} & 33.89/0.9290 & 31.23/0.8830 & {27.92/0.7896} & 32.98/0.9314 & 30.81/0.9015 & 27.59/0.8331 \\
IRCNN~\cite{zhang2017learning} & 33.87/0.9285 & 31.18/0.8824 & 27.88/0.7898 & 27.59/0.8331 & 31.20/0.9088 & 27.70/0.8396 \\
FFDNet~\cite{zhang2018ffdnet} & 33.87/0.9290 & 31.21/0.8821 & 27.96/0.7887 & 33.83/0.9418 & 31.40/0.9120 & 28.05/0.8476 \\
BRDNet~\cite{tian2020image} & 34.10/0.9291 & 31.43/0.8847 & 28.16/0.7942 & 34.42/0.9462 & 31.99/0.9194 & 28.56/0.8577 \\
AirNet~\cite{li2022all} & 34.14/\underline{0.9356} & 31.48/0.8928 & 28.23/\underline{0.8057} & 34.40/0.9487 & 32.10/0.9240 & 28.88/\underline{0.8702} \\
MPRNet~\cite{zamir2021multi} & 34.09/0.9313 & 31.38/0.8856 & 28.07/0.7967 & 34.40/0.9463 & 31.91/0.9166 & 28.52/0.8576 \\
Restormer~\cite{zamir2022restormer} & \underline{34.24}/0.9337 & \underline{31.57/0.8898} & \underline{28.26}/0.8026 & \underline{34.71/0.9490} & \underline{32.30/0.9231} & \underline{29.02}/0.8685 \\
Ours & \textbf{34.30/0.9356} & \textbf{31.65/0.8916} & \textbf{28.38/0.8065} & \textbf{34.80/0.9502} & \textbf{32.48/0.9263} & \textbf{29.31/0.8774} \\\hline
\end{tabular}%
% }
\end{table*}

\begin{table*}[t]
\centering
\vspace{-0.2cm}
\caption{Quantitative results of image deraining on the Rain100L~\cite{yang2019joint} dataset.}
\label{tab:4deraining}
\vspace{-0.1cm}
% \resizebox{\textwidth}{!}{%
\begin{tabular}{c|cccccccccc}
\hline
Metrics & DIDMDN~\cite{zhang2018density} & UMRL~
\cite{yasarla2019uncertainty} & SIRR~\cite{wei2019semi} & MSPFN~\cite{jiang2020multi} & LPNet~\cite{fu2019lightweight} & AirNet~\cite{li2022all} & DRT~\cite{liang2022drt} &Restormer~\cite{zamir2022restormer} & Ours \\ \hline
{PSNR} & 23.79 & 32.39 & 32.37 & {33.50} & {33.61} & {34.90} &37.65 & \underline{38.05} & \textbf{38.33} \\
SSIM & 0.7731 & 0.9210 & 0.9258 & 0.9480 & 0.9583 & {0.9660} & 0.9750& \underline{0.9798} & \textbf{0.9839} \\ \hline
\end{tabular}%
% }
\end{table*}

\begin{table*}[t]
\centering
\vspace{-0.2cm}
\caption{Quantitative results of dehazing on the SOTS~\cite{li2018benchmarking} dataset. }
\label{tab:4dehazing}
\vspace{-0.1cm}
\begin{tabular}{c|cccccccccc}
\hline
Metrics & MSCNN~\cite{ren2016single} & AODNet~\cite{li2017aod} & EPDN~\cite{qu2019enhanced} & FDGAN~\cite{dong2020fd} & AirNet~\cite{li2022all} & DehazeFormer~\cite{song2023vision} & Restormer~\cite{zamir2022restormer} & Ours \\ \hline
{PSNR} &  22.06 & 20.29 & 22.57 & 23.15 & {23.18} & 30.87 & \underline{31.47} & \textbf{31.96} \\
SSIM &  0.9078 & 0.8765 & 0.8630 & {0.9207} & 0.9000 & 0.9755 & \underline{0.9785} & \textbf{0.9804}\\ \hline
\end{tabular}%
% }
\end{table*}

\begin{table*}[t]
\centering
\vspace{-0.2cm}
\caption{Quantitative results of image SR on the DIV2K~\cite{Agustsson_2017_CVPR_Workshops} dataset.}
\label{tab:4sr}
\vspace{-0.1cm}
% \resizebox{\textwidth}{!}{%
\begin{tabular}{c|cccccccc}
\hline
Metrics & 
EDSR~\cite{lim2017enhanced} &
CSNLN~\cite{mei2020image} &
RCAN~\cite{zhang2018residual} & HAN+~\cite{niu2020single} & 
AirNet~\cite{li2022all} & Restormer~\cite{zamir2022restormer} &  Ours \\\hline
{PSNR} & 33.15 & 33.35 & 33.43 & 33.46  & 33.52 & \underline{33.57}& \textbf{33.64} \\
SSIM & 0.9243 & 0.9274 & 0.9275 & 0.9276 & 0.9279 & \underline{0.9299} & \textbf{0.9301} \\ \hline
\end{tabular}
% }
\vspace{-0.2cm}
\end{table*}

\vspace{-0.4cm}
\subsection{Experiments on Single Degradation}\label{sec:4A}
\paragraph{Experimental settings}

We conduct experiments on four types of degradations, including noise, rain, haze, and spatial downsampling, where corresponding restoration tasks are denoising, deraining, dehazing, and SR. For denoising, we take the widely-used BSD400~\cite{martin2001database} and WEB~\cite{ma2016waterloo} datasets as training sets, and the BSD68~\cite{martin2001database} and Urban100~\cite{huang2015single} datasets as testing sets. To be detailed, BSD400~\cite{martin2001database}, WED~\cite{ma2016waterloo}, BSD68~\cite{martin2001database}, and Urban100~\cite{huang2015single} datasets consist of 400,~4,744,~68, and 100 clean
natural images, respectively. Following~\cite{li2022all,tian2020image,zhang2017beyond,zhang2017learning,zhang2018ffdnet}, Gaussian noise is added with standard deviation values of 15, 25, and 50 on the clean images. For deraining, we use the Rain100L~\cite{yang2019joint} dataset for training and testing. The training set contains 200 synthetic rainy images and their clean counterparts, and the testing set contains 100 rainy-clean image pairs. For dehazing, we use the RESIDE~\cite{li2018benchmarking} dataset, which consists of the Outdoor Training Set (OTS) and the Synthetic Objective Testing Set (SOTS) for training and testing, respectively. The OTS contains 72,135 outdoor hazy-clean image pairs and the SOTS contains 500 outdoor hazy-clean image pairs. For SR, we use the DIV2K~\cite{Agustsson_2017_CVPR_Workshops} dataset to generate low-resolution images with scaling factors of 2, where the first 750 images are used for training and the rest 50 images are used for testing. The corresponding high-resolution images are used as the ground truth. {Each image's degradation is randomly generated according to physical degradation processes~\cite{li2018benchmarking,yang2019joint,batson2019noise2self}, resulting in different degradation patterns.}

We select several recent methods as a baseline for comparison. For denoising, we compare our methods with CBM3D~\cite{dabov2007color}, DnCNN~\cite{zhang2017beyond}, IRCNN~\cite{zhang2017learning}, FFDNet~\cite{zhang2018ffdnet}, and BRDNet~\cite{tian2020image}. 
For deraining, we compare our methods with DIDMDN~\cite{zhang2018density}, UMRL~\cite{yasarla2019uncertainty}, SIRR~\cite{wei2019semi}, MSPFN~\cite{jiang2020multi}, and LPNet~\cite{fu2019lightweight}. For dehazing, we compare our methods with DehazeNet~\cite{cai2016dehazenet}, MSCNN~\cite{ren2016single}, AODNet~\cite{li2017aod}, EPDN~\cite{qu2019enhanced}, FDGAN~\cite{dong2020fd}, and DehazeFormer~\cite{song2023vision}. For SR, we compare our method with SR methods including EDSR~\cite{lim2017enhanced}, RCAN~\cite{zhang2018residual}, HAN+~\cite{niu2020single}, SwinIR~\cite{liang2021swinir}, HAT~\cite{chen2023activating}, and Restormer~\cite{zamir2022restormer}. We also compare with image restoration methods including AirNet~\cite{li2022all}, MPRNet~\cite{zamir2021multi}, and Restormer~\cite{zamir2022restormer}.

All the methods are trained and tested using the same training and testing sets. 
We trained our model using the Adam optimizer with a learning rate of $10^{-4}$. The batch size is set to $4$ and the images are cropped to size of $128\times 128$. We use a weight decay of $10^{-4}$ and a momentum of $0.9$. All the experiments are conducted on a single NVIDIA GTX 3090Ti GPU in PyTorch. We follow the same evaluation metrics as in previous works~\cite{li2022all}, including Peak Signal-to-Noise Ratio (PSNR) and structural similarity index measure (SSIM).

\begin{table*}[!ht]
\centering
\caption{Quantitative results for all-in-one image restoration.}
\label{tab:4allinone1}
\vspace{-0.2cm}
\renewcommand{\arraystretch}{1.1} 
% \resizebox{\textwidth}{!}{%
\begin{tabular}{c|c|ccc|c|c|c}
\hline
\multirow{2}{*}{Training datasets}
 & \multirow{2}{*}{Methods}   &  \multicolumn{3}{c|}{Noise}  & {Rain} & {Haze} & \multirow{2}{*}{Average}   \\ \cline{3-7}
 & & BSD68 ($\sigma$=15) & BSD68 ($\sigma$=25) &         BSD68 ($\sigma$=50) & Rain100L & SOTS & \\ \hline

\multirow{5}{*}{Noise+Rain}
& NAFNet~\cite{chen2022simple}  & 33.85/0.9302 & 31.23/0.8853  & 27.98/0.7954 & 37.19/0.9794 & -  & 32.56/0.8975  \\
 
& MPRNet~\cite{mehri2021mprnet} & 33.87/0.9334 & 31.21/0.8845  & 27.91/0.7924  & 38.15/0.9818 & -  & 32.78/0.8980 \\

& AirNet~\cite{li2022all}       & \underline{34.11/0.9352} & \underline{31.46}/\textbf{0.8923}  & \underline{28.19}/\textbf{0.8042}   & \underline{38.31/0.9824}  & - & \underline{33.01}/\textbf{0.9035} \\

& Restormer~\cite{zamir2022restormer} & 33.97/0.9342 & {31.19/0.8824} & 28.04/0.8025 & {37.10/0.9784} & - & 32.57/0.8993 \\

& Ours   & \textbf{34.11/0.9353} & \textbf{31.48}/\underline{0.8863} & \textbf{28.21}/\underline{0.8038} & \textbf{38.34/0.9824} &  - & \textbf{33.03}/\underline{0.9019} \\
 \hline

\multirow{5}{*}{Noise+Haze} 

& NAFNet~\cite{chen2022simple} & 33.28/0.9219 & 30.73/0.8741 & 27.49/0.7732 & - & 26.01/0.9439  & 29.37/0.8782 \\

& MPRNet~\cite{mehri2021mprnet}& 32.91/0.9206 & 30.03/0.8716 & 27.00/0.7671 & - & \underline{28.15/0.9605} & 29.51/0.8799 \\

& AirNet~\cite{li2022all}  & \underline{33.77/0.9299} & \underline{31.22/0.8855}    & \underline{28.00/0.7960} & - &  27.03/0.9599  & 30.01/\underline{0.8928}\\

& Restormer~\cite{zamir2022restormer} & 33.75/0.9306 & {30.96/0.8753} & 27.83/0.7854 & - & {28.12/0.9601} & \underline{30.16}/{0.8878} \\

& Ours  & \textbf{33.99/0.9339} & \textbf{31.33/0.8891} & \textbf{28.11/0.7997} & - & \textbf{28.65/0.9642} & \textbf{30.50/0.8952}\\
 \hline

\multirow{5}{*}{Rain+Haze} 
& NAFNet~\cite{chen2022simple}   & - & - & - & 32.87/0.9471 & 26.99/0.9513 & 29.93/0.9492 \\
& MPRNet~\cite{mehri2021mprnet}  & - & - & - & 33.70/0.9550 & 26.58/0.9533 & 30.14/0.9541\\

& AirNet~\cite{li2022all}        & - & - & - & 32.50/0.9465 & 26.78/0.9577           & 29.64/0.9521 \\
& Restormer~\cite{zamir2022restormer} & - & - & - & \underline{34.58/0.9633} & \underline{27.99/0.9584} & \underline{31.28/0.9608} \\
& Ours                                & - & - & - & \textbf{35.42/0.9695}    & \textbf{28.16/0.9585}    & \textbf{31.79/0.9640} \\
 \hline

\multirow{5}{*}{\begin{tabular}[c]{@{}c@{}}Noise+\\ Rain+Haze\end{tabular}} 
& NAFNet~\cite{chen2022simple}     & 33.03/0.9176 & 30.47/0.8649  & 27.12/0.7540 & 33.64/0.9560 & 24.11/0.9275  & 29.67/0.8440 \\
& MPRNet~\cite{mehri2021mprnet}    & 33.27/0.9196 & 30.76/0.8710  & 27.29/0.7613 & 33.86/0.9579 & \underline{28.00}/0.9582  & 30.63/0.8936 \\
& AirNet~\cite{li2022all}          & \underline{33.92}/\textbf{0.9329} & \underline{31.26}/\textbf{0.8884} & \underline{28.00/0.7974} & \underline{34.90/0.9675}           & 27.94/\underline{0.9615}    & \underline{31.20/0.9095} \\
& Restormer~\cite{zamir2022restormer} & 33.72/0.9298 & {30.67/0.8649}& 27.63/0.7922 & {33.78/0.9582} & {27.78/0.9579} & 30.75/0.9010 \\
& Ours                             & \textbf{34.01}/\underline{0.9315} & \textbf{31.36}/\underline{0.8873} & \textbf{28.10/0.7984} & \textbf{35.42/0.9685} &\textbf{28.64/0.9616} & \textbf{31.51/0.9095} \\\hline

\end{tabular}%
% }
\vspace{-0.5cm}
\end{table*}

\begin{figure*}[!ht]
  \centering
  \includegraphics[width=1\linewidth]{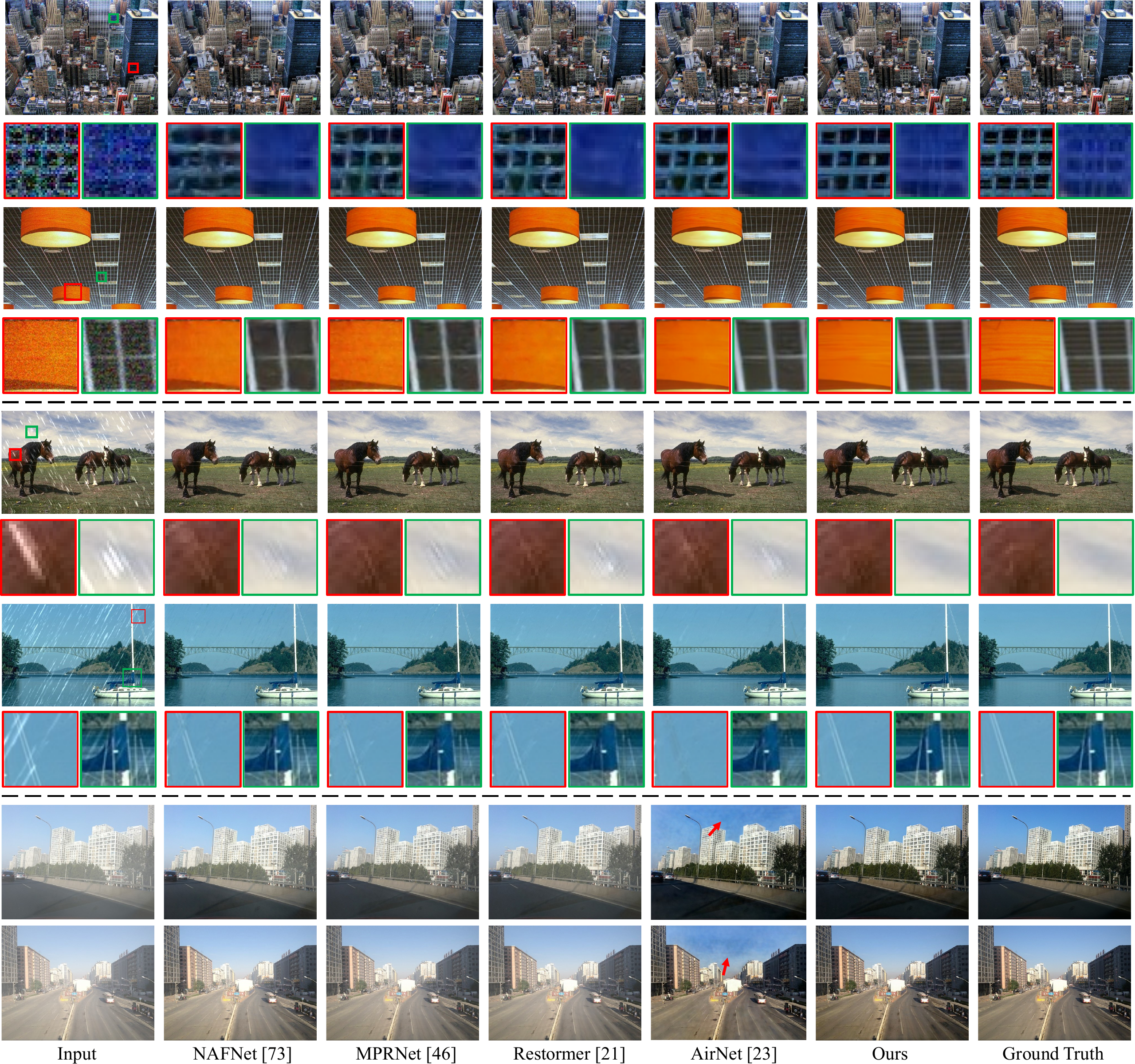}
  \vspace{-0.4cm}
  \caption{Qualitative comparisons of all-in-one image restoration. First block: Denoising on the Urban100~\cite{huang2015single} dataset ($\sigma=25$). Middle block: Deraining on the Rain100L~\cite{yang2019joint} dataset. Last block: Dehazing on the SOTS~\cite{li2018benchmarking} dataset.}
  \label{fig:4allinone}
  \vspace{-0.5cm}
\end{figure*}

\paragraph{Quantitative results}
We perform a comprehensive quantitative comparison of our method against baseline methods on various image restoration tasks. For the denoising task, as shown in Table~\ref{tab:4denoising}, our method consistently outperforms existing methods on different datasets and at various noise levels. Notably, recent image restoration methods do not specifically tailor their network architecture for denoising but exhibit better performance than previous denoising methods. For the deraining task, Table~\ref{tab:4deraining} shows the numerical results, demonstrating the robustness of our method in effectively removing rain streaks from rainy images. We also show the dehazing and SR results in Table~\ref{tab:4dehazing} and Table~\ref{tab:4sr}, respectively. These quantitative results further validate the excellent performance of our method in handling single types of degradation. All these results demonstrate the effectiveness of our method on representative restoration tasks, while the SOTA performance on single degradation is not our goal.

\vspace{-0.2cm}
\subsection{Experiments on Multiple Degradations}\label{sec:4exp_on_multi}
\subsubsection{Experimental settings}

To validate the effectiveness of our method, we conduct experiments on multiple types of image degradations, \textit{i.e.}, all-in-one image restoration. During the training phase, we mix datasets containing different types of degradations, allowing the network to learn from a variety of degradations. Subsequently, we evaluate the network's performance on multiple ``unknown'' degradations. It is worth noting that our definition of the ``unknown'' degradation aligns with~\cite{li2022all} which refers to the unspecific degradation that is not explicitly recognized, rather than degradations that are unseen during the training phase.

\begin{table*}[t]
\centering
\vspace{-0.2cm}
\caption{Quantitative results for all-in-one image restoration with $2\times$ SR.}
\label{tab:4allinone2}
\vspace{-0.1cm}
\renewcommand{\arraystretch}{1.0} 
\begin{tabular}{c|c|ccc|c|c|c}
\hline
\multirow{2}{*}{Training datasets}
 & \multirow{2}{*}{Methods} & \multicolumn{3}{c|}{Noise}  & {Rain} & {Haze} & \multirow{2}{*}{Average} \\ \cline{3-7}
 & &BSD68 ($\sigma$=15) & BSD68 ($\sigma$=25) & BSD68 ($\sigma$=50) & Rain100L & SOTS & \\ \hline

\multirow{5}{*}{Noise+Rain}
& NAFNet~\cite{chen2022simple}  & 28.24/0.8173 & 27.39/0.7781  & 25.63/0.6930  & 27.59/0.8246     & -  & 27.21/0.7782  \\
 
& MPRNet~\cite{mehri2021mprnet} & 28.64/0.8324 & 27.77/0.7932  & 26.01/0.7139  & 28.67/0.8540 & -  & 27.77/0.7983 \\

& AirNet~\cite{li2022all}  & \underline{28.71/0.8326} & \underline{27.88/0.7998} & \underline{26.28}/\textbf{0.7276} & 28.22/0.8417 & - & 27.77/\underline{0.8004} \\

& Restormer~\cite{zamir2022restormer} & 28.67/0.8322 & {27.75/0.7914} & 26.11/0.7155 & \underline{29.16/0.8593} & - & \underline{27.92}/0.7996 \\

& Ours   & \textbf{28.89/0.8341} & \textbf{28.03/0.8006} & \textbf{26.37}/\underline{0.7234} & \textbf{29.54/0.8612} &  - & \textbf{28.20/0.8048} \\
 \hline

\multirow{5}{*}{Noise+Haze} 

& NAFNet~\cite{chen2022simple} & 28.20/0.8149 & 27.33/0.7770  & 25.60/0.6932 & - & 24.46/0.8399 & 26.39/0.7812 \\

& MPRNet~\cite{mehri2021mprnet}& 28.10/0.8121 & 27.38/0.7762  & 25.68/0.6954 & - & 24.29/0.8438 & 26.36/0.7818\\

& AirNet~\cite{li2022all} & 28.46/\underline{0.8262} & 27.71/0.7922 & 26.00/0.7123 & - & 24.85/0.8471 & 26.75/0.7944 \\

& Restormer~\cite{zamir2022restormer} & \underline{28.52}/0.8236 & \underline{27.75/0.7923} & \underline{26.03/0.7142} & - & \textbf{25.15}/\underline{0.8522} & \underline{26.86/0.7954} \\

& Ours   & \textbf{28.63/0.8342} & \textbf{27.91/0.7969} & \textbf{26.22/0.7205} & - & \underline{25.13}/\textbf{0.8578} & \textbf{26.97/0.8023} \\
 \hline

\multirow{5}{*}{Rain+Haze} 
& NAFNet~\cite{chen2022simple}   & - & - & - & 27.42/0.8208 & 24.09/0.8317 & 25.75/0.8262 \\
& MPRNet~\cite{mehri2021mprnet}  & - & - & - & 27.10/0.8184 & 23.90/0.8367 & 25.50/0.8275\\
& AirNet~\cite{li2022all}        & - & - & - & 28.52/0.8447 & 24.93/0.8520 & 26.72/0.8483 \\
& Restormer~\cite{zamir2022restormer} & - & - & - & \underline{28.72/0.8499} & \underline{25.11/0.8545} & \underline{26.91/0.8522}\\
& Ours                                & - & - & - & \textbf{29.16/0.8600}    & \textbf{25.33/0.8571}    & \textbf{27.24/0.8587} \\
 \hline

\multirow{5}{*}{\begin{tabular}[c]{@{}c@{}}Noise+\\ Rain+Haze\end{tabular}} 
& NAFNet~\cite{chen2022simple}     & 28.20/0.8161 & 27.36/0.7778  & 25.63/0.6949 & 27.53/0.8230 & 24.89/0.8444  & 26.72/0.7912 \\
& MPRNet~\cite{mehri2021mprnet}    & 27.76/0.8150 & 26.93/0.7746  & 25.35/0.6878 &  27.73/0.8229 & 22.76/0.8300  & 26.10/0.7860 \\
& AirNet~\cite{li2022all}   & 28.47/\underline{0.8252} & 27.68/0.7925    & 26.05/0.7179 & \underline{27.99/0.8361} & 24.22/0.8371 & 26.88/0.8019\\
& Restormer~\cite{zamir2022restormer} & \underline{28.51}/0.8252 & \underline{27.83/0.7951} & \underline{26.11/0.7184} & 27.78/0.8358 & \underline{24.78/0.8521} & \underline{27.00/0.8053} \\
& Ours     & \textbf{28.72/0.8258} & \textbf{27.88/0.7978} & \textbf{26.18/0.7195} & \textbf{28.62/0.8477} & \textbf{25.01/0.8601} & \textbf{27.28/0.8101} \\\hline
\end{tabular}%
% }
\vspace{-0.3cm}
\end{table*}

\begin{figure*}[!t]
  \centering
  \includegraphics[width=1\linewidth]{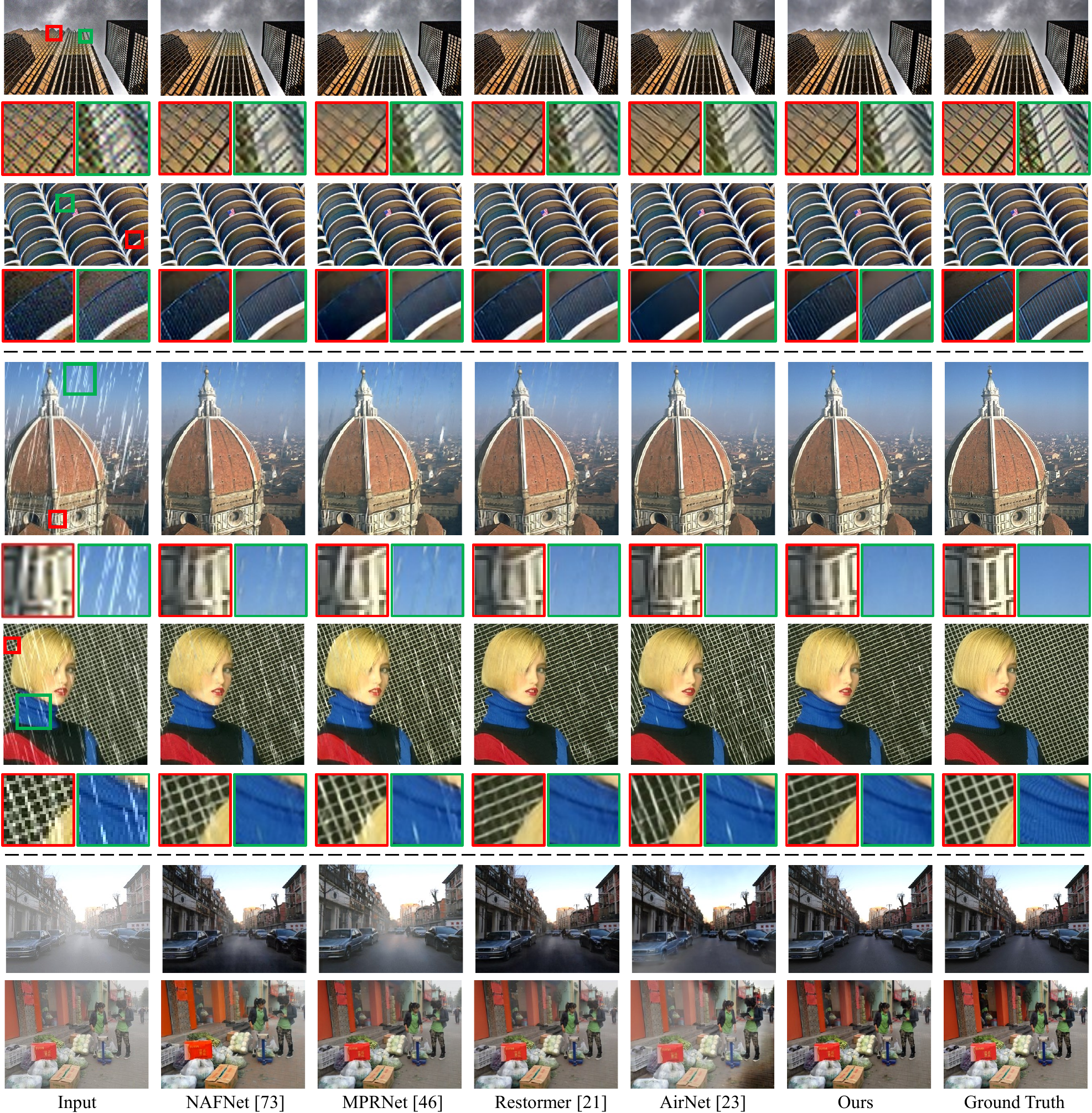}
    \vspace{-0.6cm}
  \caption{Qualitative comparisons of all-in-one image restoration with $2\times$ SR. First block: Denoising on the Urban100~\cite{huang2015single} dataset ($\sigma=25$). Middle block: Deraining on the Rain100L~\cite{yang2019joint} dataset. Last block: Dehazing on the SOTS~\cite{li2018benchmarking} dataset.}
  \label{fig:4allinonesr}
\vspace{-0.3cm}
  
\end{figure*}

We use four mixing configurations including denoising+deraining, denoising+dehazing, deraining+dehazing, and denoising+deraining+dehazing for training and evaluation. We use the same training datasets in Sec.~\ref{sec:4A} and mix them together. The datasets are resampled to balance the training data. We compare our method with various image restoration techniques, including NAFNet~\cite{chen2022simple}, MPRNet~\cite{zamir2021multi}, Restormer~\cite{zamir2022restormer}, and AirNet~\cite{li2022all}. Except for Airnet, other baselines are trained for different configurations. We only re-train AirNet for Noise+Haze since the officially reported value is extremely low, while other values follow the original paper. We also conduct experiments with scale changes to consider the SR task. PSNR and SSIM are chosen as evaluation metrics.

% \vspace{-0.1cm}
\subsubsection{Quantitative results}

We quantitatively evaluate the methods on multiple degradations at the original scale, and the results are shown in Table~\ref{tab:4allinone1}. It can be seen that, our method outperforms the baseline methods across various configurations. Notably, in the denoise+dehaze+derain setting, our dehazing performance shows an improvement of over $0.7$dB compared with AirNet~\cite{li2022all}, while the deraining performance also gains approximately $0.5$dB over AirNet. Similarly, for other configurations, significant performance gains are achieved over the baseline methods. An observation is that AirNet seems sensitive to the mixing of different degradations. In the denoise+derain configuration, it achieves a $4$dB performance change ($38.31$ vs. $33.61$) in the deraining task compared with AirNet's result in Table~\ref{tab:4deraining}. In the denoise+dehaze+derain configuration, the dehazing results have a performance gap of approximately 4dB ($27.94$ vs. $23.18$) compared with the results in Table~\ref{tab:4dehazing}. This phenomenon might indicate potential instability in handling various degradations with AirNet. In contrast, our method exhibits more stable performance in handling multiple degradations under different configurations, which demonstrates that our proposed method effectively copes with various degradation types.

We further evaluate our method on multiple degradations with the SR task, and the results are shown in Table~\ref{tab:4allinone2}. It is worth noting that handling SR tasks with varying degradations poses additional challenges as the downsampling degradation is considered. The results in Table~\ref{tab:4allinone2} show that our method outperforms the baseline methods across different degradations with the SR task. This indicates the effectiveness and versatility of our proposed approach in handling complex real-world image restoration tasks.

\subsubsection{Qualitative results}
We compare the visual results in Fig.~\ref{fig:4allinone} to demonstrate the effectiveness of our method on different restoration tasks. In each subfigure, we provide visual comparisons between our approach and several baseline methods for denoising, deraining, and dehazing tasks, where the models are trained under the configuration of denoise+derain+dehaze. For denoising, our method demonstrates superior noise reduction capabilities compared to the baseline methods. As seen in the first block of Fig.~\ref{fig:4allinone}, our method outperforms DnCNN and FFDNet, preserving finer structures and textures, making it visually pleasing.
In the deraining task, our method effectively removes rain streaks while preserving essential scene details. Fig.~\ref{fig:4allinone}'s second block shows that our method yields clearer derained images compared to AirNet. Notably, in the regions of the horse, our method successfully removes rain streaks, resulting in a more natural and visually appealing appearance. For the dehazing task (see the third block in Fig.~\ref{fig:4allinone}), our method excels in enhancing visibility and contrast, which reduces haze and improves image sharpness and color. Notably, the distant objects become more recognizable and vivid in the dehazed image than baseline methods. 
Similar phenomena can be observed in Fig.~\ref{fig:4allinonesr}. 
The visual results demonstrate the superiority of our method compared to baseline methods, proving the effectiveness of our method.

\vspace{-0.3cm}
\subsection{Experiments on Real-captured Images}

\paragraph{Experimental settings}
To further demonstrate the effectiveness of our method, we evaluate our method on real captured images. For noisy images, we conduct experiments using the SIDD~\cite{abdelhamed2018high} dataset, which contains 200 images captured with smartphones. For rainy images, we use the practical subset in JORDER~\cite{yang2017deep} for inference. {This subset contains 15 real-world rainy images \textit{without ground truth}. For hazy images, we use the RTTS subset in RESIDE ~\cite{li2019benchmarking} to conduct experiments, where the subset contains 4,322 hazy images collected from the Internet, \textit{without ground truth}. We compare our method with NAFNet~\cite{chen2022simple}, MPRNet~\cite{mehri2021mprnet}, AirNet~\cite{li2022all} and Restormer~\cite{zamir2022restormer}. All models are trained under the same configuration using synthetic datasets, \textit{i.e.}, noise+rain+haze, as described in Sec~\ref{sec:4exp_on_multi}.}

\begin{figure*}[t]
    \centering
    \includegraphics[width=0.98\linewidth]{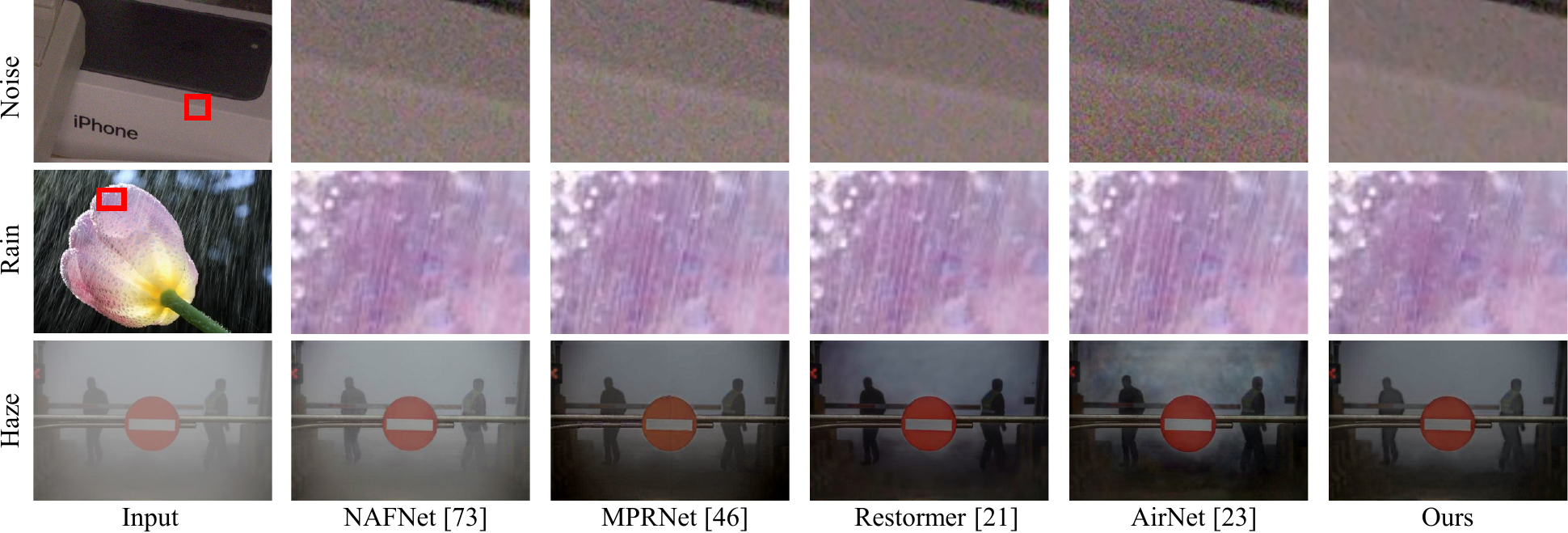}
    \caption{{Qualitative results of different methods on real-captured images. It can be seen that our method has better performance than other methods. However, there is still space for performance improvement due to the large domain gap between real-world and simulated degradation.}}
    \label{fig:4real}
    \vspace{-0.3cm}
\end{figure*}

\paragraph {Qualitative results} We show the visual results in Fig.~\ref{fig:4real}. 
{It can be seen that due to the domain gap between the degradation of the synthetic and the real world, their performance in the real-world datasets is still poor. }
{It is worth noting that AirNet tend to maintain degradations, which might be caused by the contrastive learning to force the separation of degradation types. The contrastive learning may struggle to accurately identify and address unfamiliar degradation patterns based on existing positive/negative pairs, resulting in suboptimal outcomes on new images. This issue affects our model as well, but our method demonstrates better generalization performance due to its ability to learn degradation representations. }
Moreover, there is an interesting observation that the earlier checkpoint has a better generalization ability on real-world images, which could be further explored.

\paragraph{Quantatitive results} To further demonstrate the effectiveness of our method in real-captured images, we calculate non-reference image quality assessment metrics (DBCNN~\cite{zhang2018blind} and MUSIQ~\cite{ke2021musiq}) in Table~\ref{tab:real_value}. It can be seen that, our method has better performance than other baseline methods.

\vspace{-0.3cm}
\subsection{{Experiments on single image with multiple degradations}}

\begin{figure*}[t]
    \centering
    \includegraphics[width=0.98\textwidth]{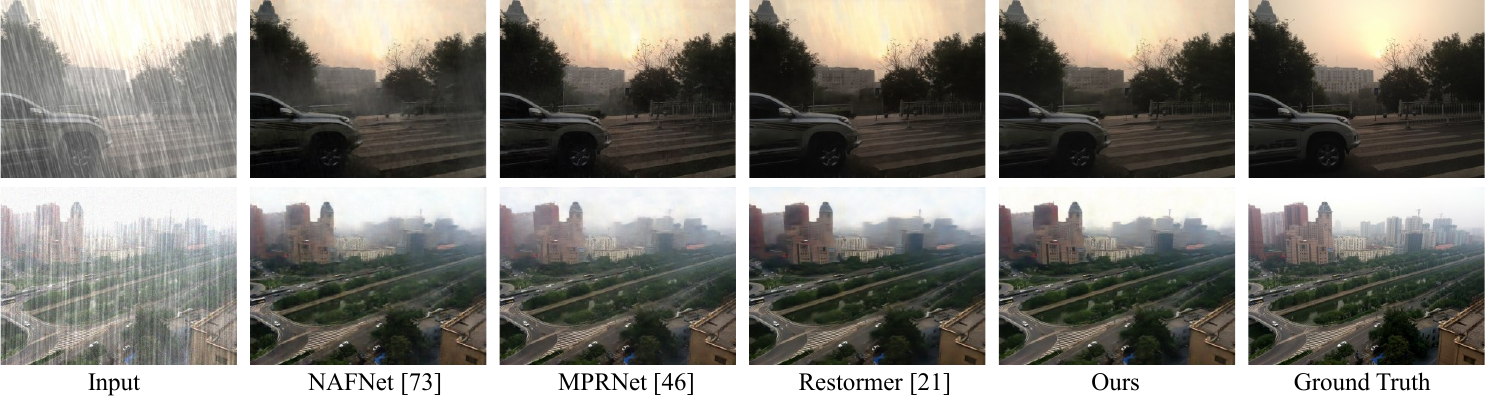}
        % \vspace{-0.2cm}
    \caption{{Qualitative results of multiple degradation types on a single image.}}
    \label{fig:multi-degradation}
    \vspace{-0.3cm}
\end{figure*}

{We demonstrate our method's effectiveness on the single image with multiple degradation modalities. First, we construct dataset by sequentially adding rain and noise degradations to the existing foggy images in the SOTS dataset~\cite{li2018benchmarking}. The constructed dataset includes 72,135 training images and 500 test images. We choose NAFNet~\cite{chen2022simple}, MPRNet~\cite{mehri2021mprnet} and Restormer~\cite{zamir2022restormer} as baselines. The quantitative results are shown in Table~\ref{tab:multi-degradation}. Our method outperforms the baselines, demonstrating its superiority. Additionally, we show the qualitative results in Fig.~\ref{fig:multi-degradation}. Despite the complexity of mixed degradations, our method still removes the degradations and restores clean images. }

\begin{table}[t]
\vspace{-0.2cm}
        \caption{Quantitative results on real-captured images in terms of non-reference metrics (DBCNN~\cite{zhang2018blind} and MUSIQ~\cite{ke2021musiq}). Higher values indicate better performance.}
    \label{tab:real_value}
    \vspace{-0.1cm}
\renewcommand{\arraystretch}{1.0} 
    \centering
    \begin{tabular}{cccccc}
    \hline
        & Noise & Rain & Haze \\\hline
        Restormer~\cite{zamir2022restormer} & \underline{27.97/30.58} & {51.62/60.35} & 42.66/\underline{53.65} \\
        AirNet~\cite{li2022all} & 26.97/29.29& \underline{52.37/60.42} & \underline{43.12}/53.57\\
        Ours & \textbf{35.04/31.97} &\textbf{52.56/60.58} & \textbf{43.77/54.24}\\\hline
    \end{tabular}
\vspace{-0.5cm}
\end{table}

\begin{table}[t]
    \centering
        \caption{{Quantitative results of multiple degradation types on a single image.}}
    \label{tab:multi-degradation}
    \vspace{-0.1cm}
    \renewcommand{\arraystretch}{1.0}
    \begin{tabular}{cccccccc}
    \hline
        {Method} & {NAFNet~\cite{chen2022simple}} & {MPRNet~\cite{mehri2021mprnet}} & {Restormer~\cite{zamir2022restormer}} & {Ours}  \\ \hline
       {PSNR}  & {24.20} & {\underline{25.65}} & {25.46} & {\textbf{26.02}} \\
       {SSIM} & {0.8169} & {0.8509} & {\underline{0.8558}} & {\textbf{0.8657}} \\
\hline
    \end{tabular}
\vspace{-0.4cm}
\end{table}

\vspace{-0.2cm}
\section{Discussions and Analyses}
\vspace{-0.1cm}

\subsection{ {Degradation Query and 1$\times$1 Conv}}\label{sec:4_11conv}
{During the two stages of the DQ module (\emph{i.e.,} Eq.~\ref{eq:get_sigma} and Eq.~\ref{eq:get_U}), we conduct matrix multiplications similar to the weighted summation process of $1\times1$ convolution. Hence, from an equivalence perspective, both stages can be implemented using $1\times 1$ convolution since they involve pixel-wise weighted summations. However, this may lead to potential misunderstandings because: 1) while both stages use NDR as the convolution kernel, the kernels in the first and second stages are transposes of each other, which is a practice rarely seen in classical $1\times 1$ convolution operations; 2) interpreting these stages solely from the perspective of $1\times 1$ convolution may diminish the physical significance of NDR. In other words, the first stage calculates the similarity using NDR and features, while the second stage resamples NDR.
}

\subsection{{Approximated degradation $U$}}
\begin{figure}[t]
    \centering    \includegraphics[width=0.48\textwidth]{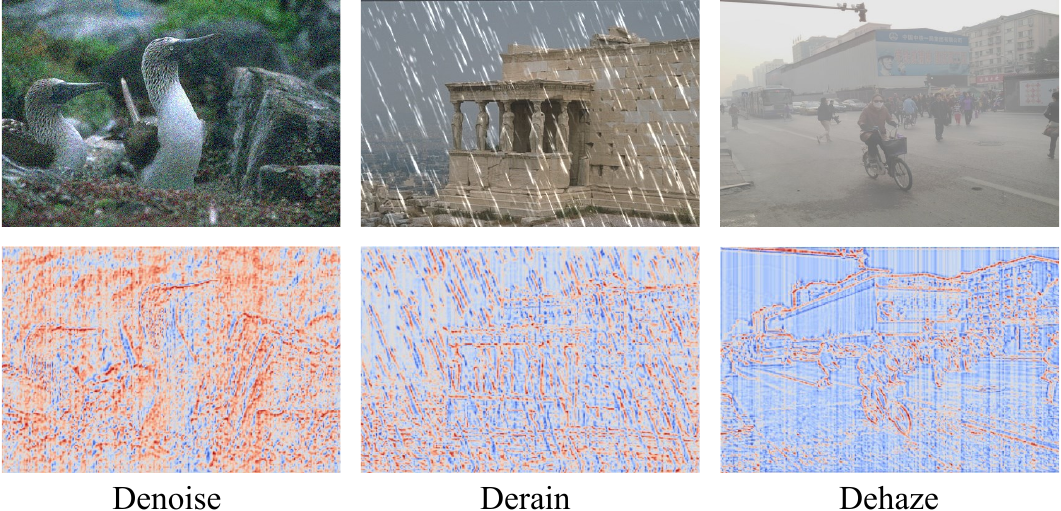}
    \caption{{Visualization of intermediate features in the encoder. The features focus on the edges and textures, instead of the degradation, demonstrating the effectiveness of NDR.}}
    \label{fig:r1_q6}
    \vspace{-0.3cm}
\end{figure}

{To demonstrate the approximated degradation $U$ could represent the degradation,  we visualize the image features in the encoder before they pass through the DQ and DI modules in Fig.~\ref{fig:r1_q6}. 
The visualizations show that the extracted features focus more on the edges and textures of the image's local areas, which are much different from the approximated degradation shown in Fig. 5.
This phenomenon aligns with how convolutional neural networks emphasize local regions and features like edges and textures~\cite{he2019modulating,peng2021conformer}.}

\begin{table}[t]
    \centering
    % \vspace{-0.2cm}
        \caption{Ablation study on NDR. $M$ is the feature dimension of degradation and $N$ is the number of degradation types. ``-'' means the network without NDR.}
    \label{tab:aba_ndr}
    \vspace{-0.1cm}
    \renewcommand{\arraystretch}{1.0}
    \begin{tabular}{ccccccc}
    \hline
        size ($M,N$)& Noise & Rain & Haze\\ \hline
        - & 30.82/0.8762 & 34.82/0.9611 & 27.83/0.9587 \\
        (32, 8) & 31.05/0.8840 & 35.27/0.9677 & 28.43/0.9603  \\ 
        (64, 32)  & 31.25/0.8862 & 35.33/0.9679 & 28.47/0.9602 \\ 
        (128, 16)  & {31.30/0.8872} & {35.37/0.9683} & {28.50/0.9609} \\ 
        (128, 32) & \textbf{31.36/0.8873} & \textbf{35.42/0.9685} & \textbf{28.64/0.9616} \\ \hline
    \end{tabular}
\vspace{-0.2cm}
\end{table}

\subsection{Ablation on NDR}
NDR plays a crucial role in capturing essential degradation characteristics, and its size might influence restoration performance. We explore the influence of NDR's size, \textit{i.e.}, feature dimensions ($M$) and degradation types ($N$), on restoration performance. We follow the  denoise+derain+dehaze configuration to train the model and test it on the BSD68 dataset. As shown in Table~\ref{tab:aba_ndr}, when we remove NDR (-), the restoration performance drops significantly, demonstrating its importance in the network. As we increase the size of NDR by expanding $M$ and $N$, we observe improved restoration performance. For example, when we increase the feature dimension from $32$ to $128$ and the number of degradation types from $8$ to $16$, the PSNR and SSIM scores show notable improvements, demonstrating the effectiveness of larger NDR sizes. {While increasing size improved restoration performance, the increase of performance shows a marginal effect beyond 16.} Therefore, we select ($128,32$) as the NDR size since it achieves a balance between representation capacity and size. {To assess the computational time effect, we increase the number of degradation types from 8 to 32 on patch sizes of $64\times64$, and observe the negligible impact on computation time (less than $10^{-4}$ seconds). }

\begin{figure}[t]
    \centering
    \includegraphics[width=0.48\textwidth]{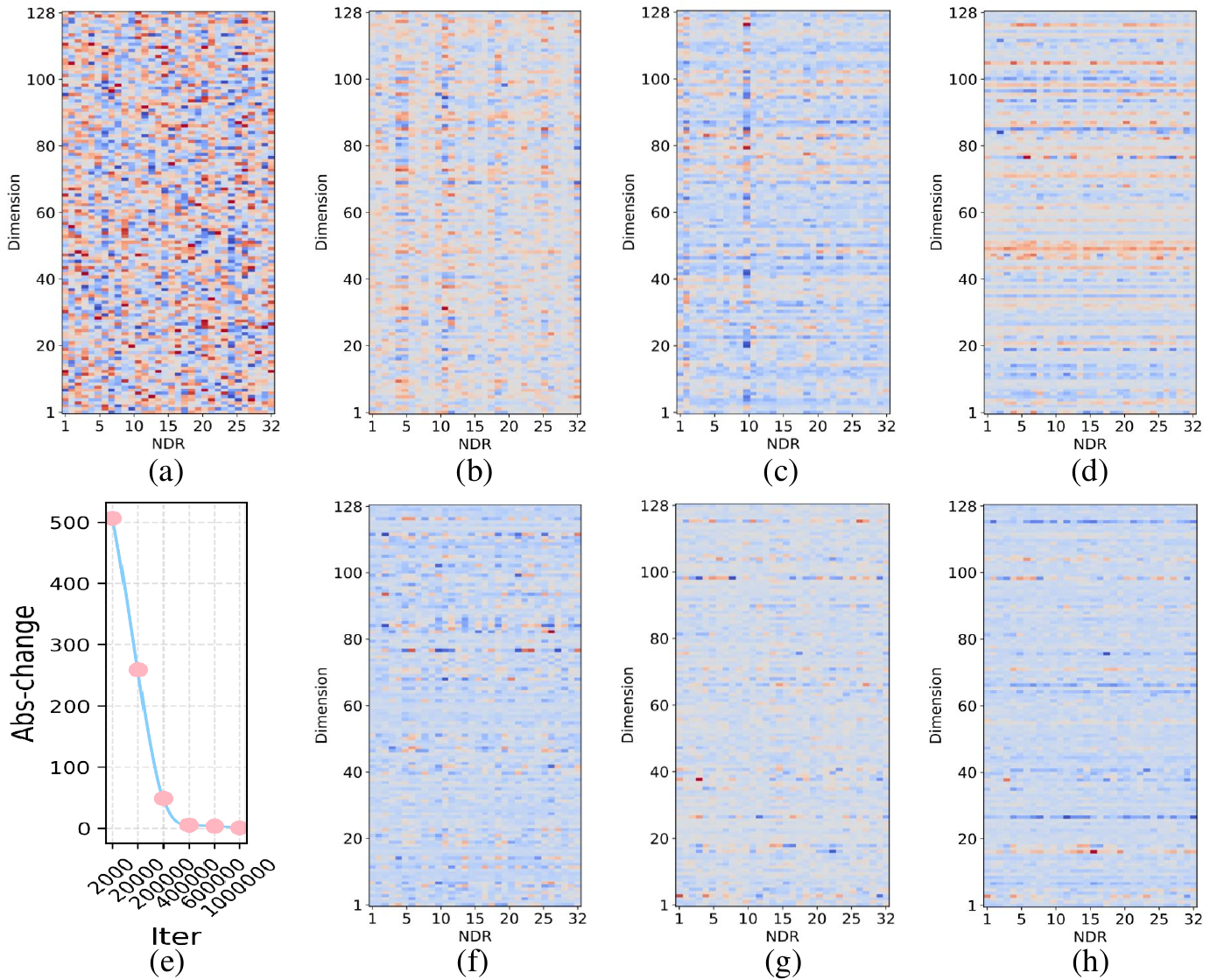}
        \vspace{-0.2cm}
    \caption{{Visualization of NDR changes throughout the optimization process. (a) The randomly initialized NDR. (b)-(d) and (f)-(h) show the changes in NDR during the optimization process, since directly visualizing the NDR appears similar. (e) The change ratio (absolute error between the current NDR and the previous one) during the optimization process.}}
    \label{fig:r3_q1}
    \vspace{-0.3cm}
\end{figure}

{We illustrate the changes of NDR in Fig.~\ref{fig:r3_q1}. Initially, NDR is completely random (Fig.~\ref{fig:r3_q1}(a)), and it gradually converges as the model trains. In Fig.~\ref{fig:r3_q1} (b)-(d) and (f)-(h), we show error maps of absolute changes between iterations. The interval for calculating error maps is shown in Fig.~\ref{fig:r3_q1}(e). As training progresses, the changes in NDR decrease, indicating it evolves from disorder to order.
}

\begin{table}[t]
    \centering
        \caption{Ablation study on the proposed modules.}
    \label{tab:aba_modules}
    \vspace{-0.1cm}
    \renewcommand{\arraystretch}{1.0} 
    \setlength\tabcolsep{4pt}
    \begin{tabular}{ccccccc}
    \hline
         & Noise & Rain & Haze \\ \hline
        w/o DQ & 30.71/0.8731 & 35.18/0.9661 & 28.15/0.9577 \\
        w/o DI & 30.75/0.8213 & 35.26/0.9673 & 27.99/0.9565\\
         w/o CP  & 31.07/0.8842 & 35.35/0.9677 & 28.34/0.9594\\ 
        Ours & \textbf{31.36/0.8873} & \textbf{35.42/0.9685} & \textbf{28.64/0.9616} \\ \hline
    \end{tabular}
\vspace{-0.1cm}
\end{table}

\begin{table}[t]
    \centering
        \caption{{Ablation study on the $\lambda$.}}
    \label{tab:aba_strategy}
    \vspace{-0.1cm}
    \renewcommand{\arraystretch}{1.0}
    \begin{tabular}{ccccccc}
    \hline
        {$\lambda$} & {Noise} & {Rain} & {Haze}  \\ \hline
       {0}  & {31.07/0.8854} & {35.20/0.9662} & {28.42/0.9581} \\
       {0.5}  & {31.24/0.8858} & {35.45/0.9688} & {28.37/0.9577} \\
        {1}  & {31.36/0.8873} & {35.42/0.9685} & {28.64/0.9616} \\ 
       {1.5}  & {31.38/0.8873} & {35.39/0.9679} & {28.39/0.9581} \\
       {3}  & {31.33/0.8862} & {35.20/0.9662} & {28.39/0.9583} \\
\hline
    \end{tabular}
\vspace{-0.3cm}
\end{table}

\begin{figure}[!t]
    \centering
    \includegraphics[width=0.98\linewidth]{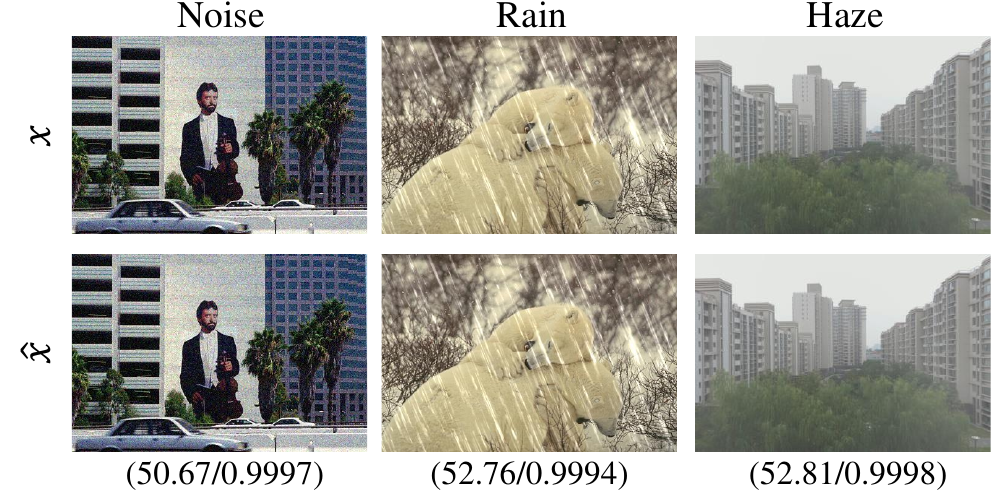}
    \caption{Visualization of NDR-Degrad's outputs. The output $\hat{x}$ exhibits similar degradations as the original degraded image $x$, demonstrating the effectiveness of NDR-Degrad in generating degraded images. PSNR and SSIM values are provided to further emphasize the high similarity between $\hat{x}$ and $x$.}  
    \label{fig:4degrad}
    \vspace{-0.3cm}
\end{figure}

\vspace{-0.2cm}
\subsection{Ablation on the Modules}
In this section, we conduct ablation experiments to evaluate the impact of our proposed DQ and DI modules. Specifically, we compare our full model by replacing the DQ and DI modules with vanilla convolution. Moreover, we conduct ablation on the CP-Conv by replacing it with concatenation \& convolution. Table~\ref{tab:aba_modules} shows the quantitative results of the ablation study, in terms of PSNR and SSIM. We follow the denoise+derain+dehaze configuration to train models and test them on BSD68~\cite{martin2001database} ($\sigma=25$), Rain100L~\cite{yang2019joint}, and SOTS~\cite{li2018benchmarking}.

We observe that removing the DQ module leads to a decrease in PSNR and SSIM values, indicating its effectiveness in approximating degradation information. Similarly, excluding the DI module also results in a notable drop, showing that the DI module plays a crucial role in utilizing the degradation information. Moreover, when CP-Conv is removed, the PSNR and the SSIM drops, indicating the necessity of CP-Conv.

\vspace{-0.2cm}
\subsection{Training Strategy}
{We demonstrate the effectiveness of our proposed training strategy by adjusting the $\lambda$ value in loss function (Eq.~\ref{eq:3bidirectional}). This study is performed on the noise+rain+haze configuration, as detailed in Sec~\ref{sec:4exp_on_multi}, and the results are presented in Table~\ref{tab:aba_strategy}. 
The results show a clear trend: when \(\lambda\) is too large (or too small), the network overemphasizes (or overlooks) modeling the degradation process. This causes the network to allocate too many (or too few) resources to capturing degradation characteristics, thereby neglecting the restoration process. }

{Furthermore, we found that due to the complexity of multiple degradation types, changes in $\lambda$ show different tendencies for each task. For example, the denoising, deraining, and dehazing tasks achieve the best results at $\lambda = 1.5$, $0.5$, and $1$, respectively. This phenomenon may be related to the coupling between different degradations, known as task dependency and conflict in multi-task learning~\cite{vandenhende2021multi}, which we will explore in future work.}

We visualize the outputs of NDR-Degrad to further emphasize the effectiveness of NDR-Degrad, which can effectively generate degraded images. As shown in Fig.~\ref{fig:4degrad}, the output image, denoted as $\hat{x}$, has a similar degradation as the original degraded image $x$, indicating that NDR-Degrad is able to degrade the image. These visualizations provide clear evidence of NDR-Degrad's efficacy, which drives NDR to learn appropriate degradation representations.

\subsection{Runtime and Parameter}

We present the parameter and runtime of our models. For NDR-Restore, it takes only 0.061 seconds to restore a $64\times64$ image on a single GTX 3090Ti GPU, with $28.4$M parameters. For NDR-Degrad, we adopt a lighter architecture with $10.70$M parameters to assist the NDR learning. Notably, our NDR module consists of just $0.004$M parameters, while the DQ and DI modules contribute $2.38$ million parameters, which incurs a relatively small overhead compared to the encoder and decoder (approximately $26.1$M parameters). This exemplifies the lightweight nature of our core design, demonstrating its potential for integration into existing network architectures.

{We also give an explicit comparison with AirNet, which has 7.61M parameters and a computational time is 0.037 seconds. These results indicate that AirNet is lightweight and fast. However, it is important to note that AirNet employs contrastive learning, which might force classifications on unseen degradations during testing, potentially reducing performance, as shown in Fig.~\ref{fig:4real}. Additionally, both AirNet and our method allow flexible backbone structure choices, which significantly affect the number of parameters and computational time. We believe exploring lighter backbone structures will be a potential direction for our future work.}

\subsection{{Approximated degradation of spatial downsampling}}

\begin{figure}[t]
    \centering
    \includegraphics[width=0.48\textwidth]{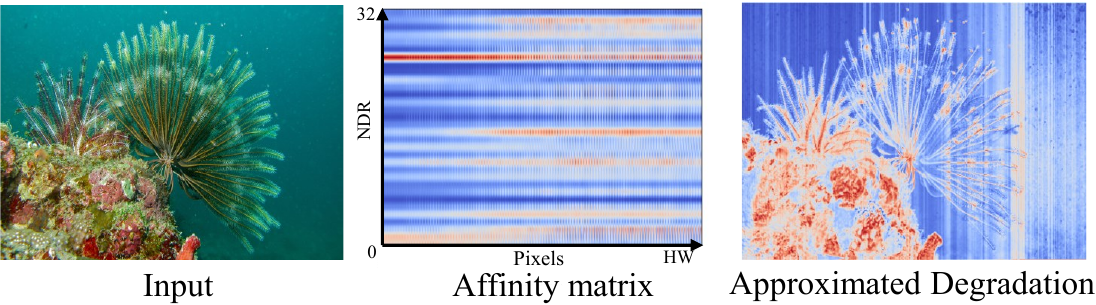}
        \vspace{-0.2cm}
    \caption{{Affinity matrix and approximated degradation of spatial downsampling.}}
    \label{fig:sr_vis}
    \vspace{-0.3cm}
\end{figure}

{We show the affinity matrix and approximated degradation of downsampling in Fig.~\ref{fig:sr_vis}. The approximated degradation tends to concentrate more on edge regions, consistent with the observation that downsampling often results in the loss of fine details. These findings further validate the effectiveness of our approach. 
However, it's important to note that this visualization only demonstrates how our method approximates and addresses degradation effects within our study's scope, not representing how downsampling degradation would manifest in the physical world.  On the other hand, visualizing downsampling degradation is generally challenging. This is because downsampling alters not only pixel distribution but also image resolution, making direct comparison between high-resolution and low-resolution images challenging.}

\begin{table}[t]
    \centering
        \caption{{Discussion on additional constraints $\mathcal{L}_e$.}}
    \label{tab:loss_entropy}
    \vspace{-0.1cm}
    \renewcommand{\arraystretch}{1.0}
    \begin{tabular}{cccccccc}
    \hline
         & {Noise} & {Rain} & {Haze} &  \\ \hline
       {w/o $\mathcal{L}_e$} & {31.36/0.8873} & {35.42/0.9662} & {28.64/0.9516} \\
        {w/ $\mathcal{L}_e$}  & {31.34/0.8869} & {35.41/0.9658} & {29.13/0.9572} &  \\
\hline
    \end{tabular}
\vspace{-0.3cm}
\end{table}

\subsection{{Additional mutual information constraint $\mathcal{L}_e$}}

We conduct experiments with an additional cross-entropy (CE) loss function $\mathcal{L}_e$ to minimize the mutual information of NDR on dimension $N$. {As shown in Table~\ref{tab:loss_entropy},  CE loss improves dehazing performance while only introducing negligible performance drops in denoising and deraining tasks, showing its potential as a promising direction to dig into.}

\subsection{{Fidelity v.s. perception}}

We explore fidelity- and perception-oriented all-in-one image restoration by comparing our approach with recent methods~\cite{potlapalli2023promptir,chen2023hat,lin2023diffbir,wang2023exploiting}.
{For fidelity-oriented methods, we follow the training configuration of PromptIR~\cite{potlapalli2023promptir} to train both HAT~\cite{chen2023hat} and our model, and we directly take the results for PromptIR from the officially published version. The results, shown in Table ~\ref{tab:promptir-hat}, demonstrate that our model still performs best in denoising, deraining, and dehazing tasks. Additionally, we notice that the data proportion for different tasks greatly affects the model's performance, which means increasing data for a specific task improves its performance but may reduce performance in other tasks~\cite{xu2023learning,jiang2021improving}.}

% compared the PSNR and SSIM metrics. As shown in Table~\ref{tab:promptir-hat}, our results generally outperform the two baselines in denoising and dehazing tasks. To ensure a fair comparison, we re-train both networks using our experimental setup.}

\begin{figure}[t]
    \centering
    \includegraphics[width=0.48\textwidth]{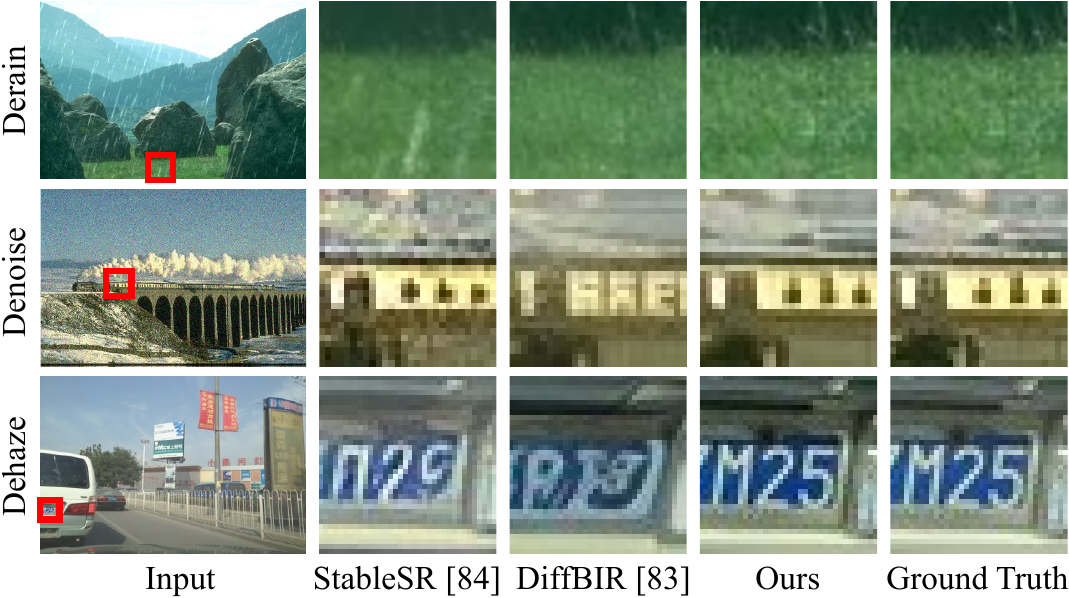}
        % \vspace{-0.2cm}
    \caption{{Qualitative comparison with recent generative-oriented methods.}}
    \label{fig:diffbir}
    \vspace{-0.3cm}
\end{figure}

{For perception-oriented methods (\textit{i.e.,} DiffBIR~\cite{lin2023diffbir} and StableIR~\cite{wang2023exploiting}), we focus on visual quality, as shown in Fig.~\ref{fig:diffbir}. It can be observed that generative methods, leveraging priors from existing diffusion models, tend to produce text/texture hallucinations. In contrast, our method shows a stable and accurate restoration performance on multiple tasks.}

\begin{table}[t]
    \centering
        \caption{{Quantitative comparison with recent fidelity-oriented methods.} }
    \label{tab:promptir-hat}
    \vspace{-0.1cm}
    \renewcommand{\arraystretch}{1.0}
    \begin{tabular}{cccccccc}
    \hline
        {Method} & {Noise} & {Rain} & {Haze} &  \\ \hline
        
    \hline  {HAT~\cite{chen2023hat}} & {31.05/0.879} & {32.98/0.957} & {29.46/0.973} \\
        
      {PromptIR~\cite{potlapalli2023promptir}} & {\underline{31.31}/\textbf{0.888}} & {\underline{36.37/0.972}} & {\underline{30.58/0.974}} &  \\

        {Ours}  & {\textbf{31.36}/\underline{0.886}} & {\textbf{36.41/0.975}} & {\textbf{31.03/0.977}} \\
        
\hline
    \end{tabular}
\vspace{-0.3cm}
\end{table}

\subsection{{Performance on the conflict task}}

{
To validate the model's capability to handle ``conflicting'' tasks, we conduct experiments on exposure correction using the MSEC dataset~\cite{afifi2021learning}. The dataset includes underexposed and overexposed images at five exposure levels. We use 17,675 and 5,905 image pairs for training and testing, respectively. }

\begin{figure}[t]
    \centering    \includegraphics[width=0.48\textwidth]{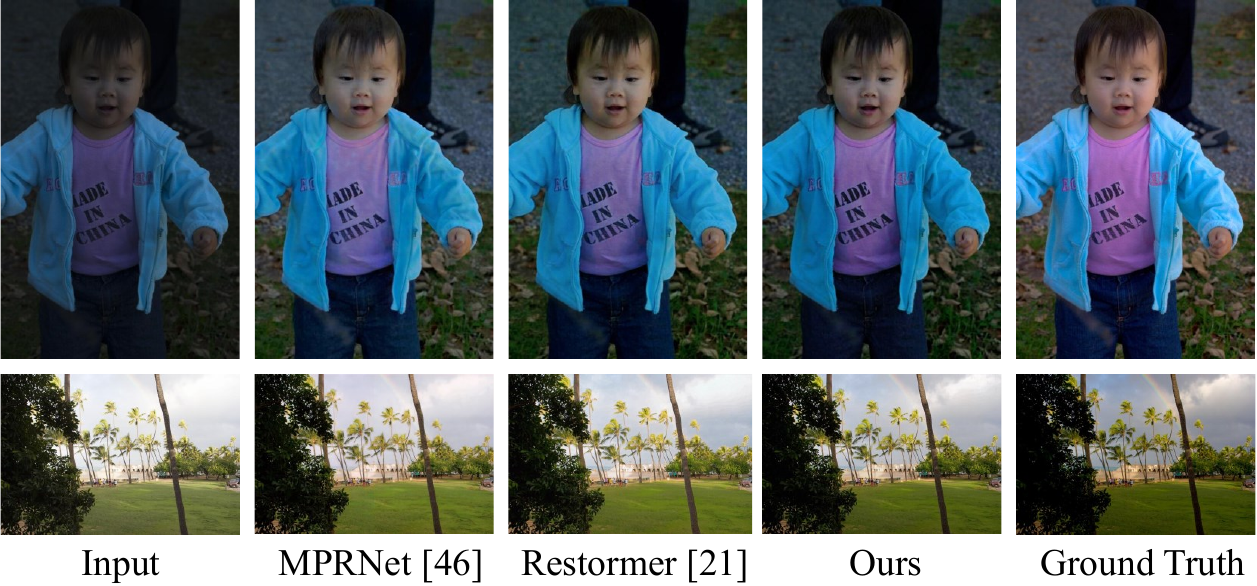}
    \caption{{Visualization results of (top) underexposure correction and (bottom) overexposure correction on the MSEC dataset~\cite{afifi2021learning}.} }
    \label{fig:exposure}
    \vspace{-0.3cm}
\end{figure}

{The quantitative results are shown in Table~\ref{tab:exposure}, demonstrating the model's performance separately on overexposed data, underexposed data, and their average. 
Despite not having a specialized structure for exposure correction, our method achieves the best numerical results, demonstrating its advantage in handling such conflicting tasks. Additionally, we provided visual comparisons in Fig.~\ref{fig:exposure}. As can be seen, for both the underexposed and overexposed images, there exist color and lightness shift problems of baselines. On the contrary, our method can simultaneously achieve color and lightness recovery while preserving the structures.}

\begin{table}[t]
    \centering
        \caption{{Quantitative results of exposure correction on the MSEC dataset~\cite{afifi2021learning}.}}
    \label{tab:exposure}
    \vspace{-0.1cm}
    \renewcommand{\arraystretch}{1.0}
    \begin{tabular}{ccccccc}
    \hline
        {Methods} & {Under-exposure} & {Over-exposure} & {Average}  \\ \hline
       {MPRNet~\cite{mehri2021mprnet}}  & {\textbf{21.87}/0.8137} & {18.74/0.8096} & {20.05/0.8145} \\
        {Restormer~\cite{zamir2022restormer}}  & {21.19/\underline{0.8344}} & {\underline{22.35/0.8558}} & {\underline{21.88/0.8472}} \\
        {Ours}  & {\underline{21.37}/\textbf{0.8366}} & {\textbf{22.68/0.8652}} & {\textbf{22.16/0.8537}} \\ 
\hline
    \end{tabular}
\vspace{-0.3cm}
\end{table}

\vspace{-0.2cm}
\section{Conclusions and Future Works}

In this paper, we propose NDR-Restore, an all-in-one image restoration method that can process multiple types of degradations in a single network. The key idea is to learn the NDR that effectively captures essential degradation characteristics. To leverage NDR, we propose two novel modules, the DQ module and the DI module, which effectively approximate and utilize image degradations, respectively. To drive NDR to represent degradations, we devise a bidirectional optimization strategy, where an auxiliary degradation network NDR-Degrad is jointly optimized with NDR-Restore. Experimental results demonstrated the superiority of NDR-Restore over existing methods in denoising, deraining, dehazing, and SR tasks.

{In addition to CE loss $\mathcal{L}_e$, we will explore other methods to further enhance the representation ability of NDR. 1) Adding contrastive loss could also maximize the distance between different degradation representations, thereby improving discrimination. 2) Adding sparse coding constraints could encourage sparsity in the intermediate variables to improve the representation efficiency. 3) Applying clustering methods could group similar degradation representations together while ensuring different groups are well-separated, further distinguishing different representations.}

{Besides, improving the model's lightweight nature and efficiency is also a potential direction. It is also interesting to explore the coupling relationship between different degradations, particularly task dependency and conflict in all-in-one learning.}

\ifCLASSOPTIONcaptionsoff
  \newpage
\fi

\bibliographystyle{IEEEtran}
\bibliography{reference}

\end{document}